\documentclass[11pt,a4paper]{article}
\usepackage[utf8]{inputenc}

\usepackage[spanish, main=english]{babel}

\usepackage{amsmath}
\usepackage{amsfonts}
\usepackage{amssymb}
\usepackage{booktabs}
\usepackage{geometry}
\usepackage{hyperref}
\usepackage{xurl}
\usepackage{graphicx}
\usepackage{array}
\usepackage{caption}
\usepackage{ragged2e}
\title{\textbf{From Seasonality to Semantics: Benchmarking a Hybrid Probabilistic Forecasting System for Roadblocks in Bolivia}}
\author{%
\begin{minipage}[t]{0.46\textwidth}
\centering
\textbf{Rodrigo Vargas Sainz} \\
\small Universidad Privada de Santa Cruz de la Sierra \\
\scriptsize \texttt{rodrigovargas@upsa.edu.bo} / \texttt{rodrivs@mit.edu}
\end{minipage}
\hfill
\begin{minipage}[t]{0.46\textwidth}
\centering
\textbf{Christian Berón Curti} \\
\small Universidad Privada de Santa Cruz de la Sierra \\
\scriptsize \texttt{cberon@mit.edu} / \texttt{cberon@gmail.com}
\end{minipage}
}
\date{}

\begin{document}

\twocolumn[{%
  \maketitle
  \begin{abstract}
    Roadblocks in Bolivia are a social conflict phenomenon with devastating economic impacts, estimated at losses equivalent to 4\% of the national Gross Domestic Product. Despite their recurrence and impact, there is a lack of local predictive systems to anticipate these events for logistical decision-making. This paper presents a hybrid probabilistic forecasting system that integrates time series decomposition (Prophet) with natural language processing (NLP) techniques applied to a six-year corpus of Bolivian news coverage. The methodology employs vector semantic embeddings and zero-shot classification models to capture signals of discursive escalation prior to the materialization of the roadblocks. Using an expanding walk-forward validation scheme applied over 1,762 days and seven forecasting horizons (H+1 to H+7), seven internal configurations and four external benchmarks were compared, including SARIMA and LightGBM\@. The results demonstrate that the hybrid configuration (Prophet + NLP, C6) consistently outperforms purely statistical models, achieving an AUC-ROC of 0.677 at H+1 and reducing the Brier Score by 10.9\% relative to the baseline temporal model (0.220 vs. 0.247), maintaining a statistically significant error reduction across all evaluated horizons ($p < 0.02$). This research validates that the integration of semantic news signals allows for the detection of social tension peaks not captured by historical inertia, providing a technical tool for risk management in critical transport corridors.
    
    \vspace{0.4cm}
    \vspace{0.6cm}
  \end{abstract}
}]

\section{Introduction}

Roadblocks in Bolivia have a historical trajectory that exceeds their original function as a disruptive sectorial negotiation mechanism. Although roadblocks have even colonial origins, more contemporary records exist. We could go back to the ``Water War'' (2000) or the ``Gas War'' (2003)~\cite{1}, historical events that established a model of mobilization capable of reorienting economic policy in the country~\cite{2}. During the governing period of the Movimiento al Socialismo (MAS), roadblocks underwent an institutionalization process that incorporated non-traditional social sectors into the political debate before official bodies~\cite{3}. However, recent literature regarding these events documents a transformation of this instrument: from forms of social dialogue, they have devolved into a resource for internal coercion within social organizations or populations, serving as a political pressure tool apparently oriented toward private objectives that interweave with collective demands~\cite{4}.

A portion of the reviewed studies from the 2020--2025 period attributes the intensification of road conflict to the internal division of the MAS between the factions identified as ``arcista'' and ``evista'' (support groups for former President Luis Arce Catacora versus support groups for former President Evo Morales Ayma), in which highways became arenas of dispute for various reasons, ranging from the control of the party acronym (MAS) to the fight over who decides the country's policies: the political party or the government~\cite{5}. This pattern of conflict intensified with the presidential inauguration of Rodrigo Paz Pereira in 2025, whose elimination of hydrocarbon subsidies—which raised the price of gasoline by 163\%—led to a wave of protests and roadblocks on a national scale~\cite{6}.

The economic and social impact of these episodes has been consistently documented in technical reports by various organizations and news articles. It is estimated that one month of continuous roadblocks generates losses equivalent to 4\% of the national Gross Domestic Product~\cite{7}. At the sectorial level, national industry reports losses of millions of dollars, while the tourism sector faces a crisis that jeopardizes thousands of jobs~\cite{8}. Social effects include up to a tripling of basic food prices due to shortages and documented cases of mortality associated with the inability to transfer patients or supply medical oxygen~\cite{1}. Political geography studies have also identified a pattern of territorial concentration of roadblock locations in municipalities with a strong foothold of specific organizations, such as in the Chapare region~\cite{9}.

The phenomenon of road blockages has prompted an unresolved jurisprudence debate regarding its legal status: while the political Constitution of the Bolivian state recognizes the right to freedom of assembly and expression, it does not support a right to block public roads. Several legislative proposals have existed and continue to exist, such as the bill proposing sanctions for the interruption of free transit, which have generated controversy between analysts who warn about the risks of criminalizing social protest and those who maintain the need to guarantee the right to free circulation for citizens not involved in the conflicts~\cite{10}.

Despite the magnitude of this economic and social impact, and the consolidation of roadblocks as a recurring event, there are no predictive systems that allow anticipating the occurrence of these events with sufficient foresight for logistical and risk management decision-making beyond intuition and news reading by interested companies or organizations. Although the literature on early conflict detection has validated the utility of textual sources in high-visibility contexts, an analytical gap remains regarding its applicability in scenarios of recurring low-intensity conflict. The case of road blockages in Bolivia, characterized by territorial specificity, presents a challenge that global approaches such as GDELT (Global Database of Events, Language, and Tone)~\cite{11} cannot resolve on their own.

The present work addresses this through the development of a hybrid probabilistic forecasting system that combines time series decomposition (capturing the cyclical structure of roadblocks caused by social conflicts) with natural language processing techniques applied to a six-year corpus of news coverage (capturing signals of discursive escalation prior to the materialization of the events). Through an ablation study of seven internal configurations (C1--C7) and four independent external benchmarks (regularized logistic regression, LightGBM, univariate SARIMA, and XGBoost with autoregressive features) evaluated under a walk-forward validation scheme over 1,762 out-of-sample days, we quantify whether the semantic contribution of the press is statistically significant and whether this gain is attributable to the proposed information architecture or to the learning algorithm employed.

The remainder of this article is organized as follows: Section 2 reviews the related literature on social conflict detection using NLP and hybrid time series forecasting. Section 3 describes the data and the proposed methodology. Section 4 presents the results of the ablation study and statistical significance tests. Section 5 discusses the findings and their limitations. Finally, Section 6 concludes with the contributions of this work and future research lines.

\section{Related Work}

The anticipation of sociopolitical instability has progressively transitioned from expert judgment-based approaches toward early warning systems backed by large volumes of unstructured data. The availability of global coverage databases like GDELT allows for the continuous monitoring of instability events from journalistic sources~\cite{12}. Natural language processing applied to this domain has evolved from keyword detection and the application of Graph Theory toward the structured extraction of events using specialized ontologies, notably the CAMEO (Conflict and Mediation Event Observations) framework~\cite{13}, adopted by operational systems like EMBERS, which are capable of generating structured alerts that specify actor, location, timing, and event type from open sources such as the X social network, blogs, and digital press~\cite{14}. These systems have reported an anticipatory capacity (lead time) over traditional news coverage in Latin American applications, though documented limitations persist~\cite{14}.

Previous work on forecasting civil unrest from social media has explored complementary approaches to structured event extraction. A group of researchers proposed a model based on the detection of activity cascades on Twitter to predict the occurrence of protests in various Latin American countries, under the assumption that the emergence of large-scale activity cascades constitutes an observable precursor to physical mobilization~\cite{15}. Weber et al.\ evaluated predictors derived from protest participation theory on Twitter data during the Egyptian Arab Spring, finding significant support only for the increase in the volume of expressions that explicitly anticipate future protest events, with no significant evidence for hypotheses related to the general volume of political activity on social networks~\cite{16}. In a different line, Hlatshwayo and Redl, using an index of hundreds of socioeconomic indicators, trained decision tree-based models to predict unrest one year in advance. They discovered that, in addition to the history of previous unrest, food price inflation and mobile phone penetration were the most robust predictors~\cite{17}.

Recent investigations have incorporated large language models (LLMs) as a contextual reasoning mechanism for conflict dynamics~\cite{18}. Similarly, the ExoLLM model transfers representations learned from the natural language domain to numerical forecasting, integrating textual descriptions of events~\cite{19}.

The time series forecasting literature applied to conflict phenomena has identified consistent advantages in hybrid architectures that combine statistical robustness with non-linear learning capability. The ARIMA-LSTM combination constitutes the dominant architecture in this regard: the ARIMA component captures the linear trend and seasonal structure of the series, while a Long Short-Term Memory (LSTM) network is trained on the residuals of the linear fit to capture non-linear dynamics and abrupt spikes in activity unexplained by the statistical component~\cite{20}. Subsequent works have expanded this principle through dynamic combination mechanisms that allow the system to adapt the relative weight of each component according to the observed volatility regime, optimizing the robustness of the final forecast~\cite{21}.

Beyond purely endogenous variables, recent models incorporate additional exogenous variables such as economic indicators (food inflation, exchange rates) and digital behavioral signals, including the use of anonymization networks like Tor as a proxy for censorship evasion in the face of imminent repression, and activity cascades on social networks like those described previously.

When analyzing current literature on social conflict forecasting, two clear trends emerge: First, a methodological transition away from univariate statistical models toward hybrid temporal decomposition architectures. Although conflict follows predictable seasonal structures, abrupt spikes in activity depend fundamentally on exogenous variables or the smoothing of seasonal peaks. Therefore, current systems focus on isolating deterministic components using robust statistical models, reserving natural language processing (NLP) to model residuals and latent intensities. Second, a geographical asymmetry: empirical studies concentrate on English-language contexts or events of international visibility, leaving a gap in environments of recurring conflict and low media visibility, as is the case in Bolivia.

There is no existing system in the current literature applied specifically to road blockages in Bolivia, nor an evaluation that quantifies the marginal value provided by semantic press representations over temporal decomposition models like Prophet~\cite{22}, nor one that compares such an architecture against independent machine learning and time series benchmarks. The present proposal addresses both gaps: it intentionally combines the interpretability of temporal decomposition with the local precision of encoder-only models and calibrated gradient boosting, and validates the contribution of the semantic component against four external baselines---regularized logistic regression (B1), LightGBM (B2), univariate SARIMA (B3), and XGBoost with autoregressive features (B4)---all evaluated under the same walk-forward scheme, with hyperparameters optimized over a tuning window separate from the final evaluation period to ensure the comparison is free of model selection bias.

\section{Data and Methodology}

\subsection{Data Sources}

For this study, two independent information sources with different time horizons were integrated. The first source is a historical archive from the Bolivian news portal which operates as a news aggregator that replicates and centralizes content produced by various national digital media outlets. This archive contains 916,110 raw news headlines (headlines, links, timestamps, and tags), spanning from 2008 to 2026. For the analysis period (2020--2026), the thematically filtered corpus comprises 386,884 articles distributed across 2,137 days, with an average of 181 relevant news items per day. The remaining 517,426 articles correspond to coverage prior to 2020, a period with no road event records in the ABC, and were discarded from the evaluation window. The second source comes from the official records of the Administradora Boliviana de Cacishared period (2020--2026), filtering the ABC data to exclusively include disruptions due to social conflicts in six strategic corridors for national transport (Routes 1, 4, 5, 6, 7, and 25). The aggregation of the target variable at the national level responds to an intrinsic constraint of the textual data source: more than 90\% of the press headlines omit the specific nomenclature of the road route (e.g., Route 4 or Route 1), limiting themselves to reporting the intentionality of the social conflict. Forcing a spatial disaggregation at the route level would introduce a missing data bias and drastically reduce the statistical power of the dataset. Therefore, the hybrid framework is formulated as a proxy for national systemic risk.

\begin{figure*}[t]
\centering
\includegraphics[width=\textwidth]{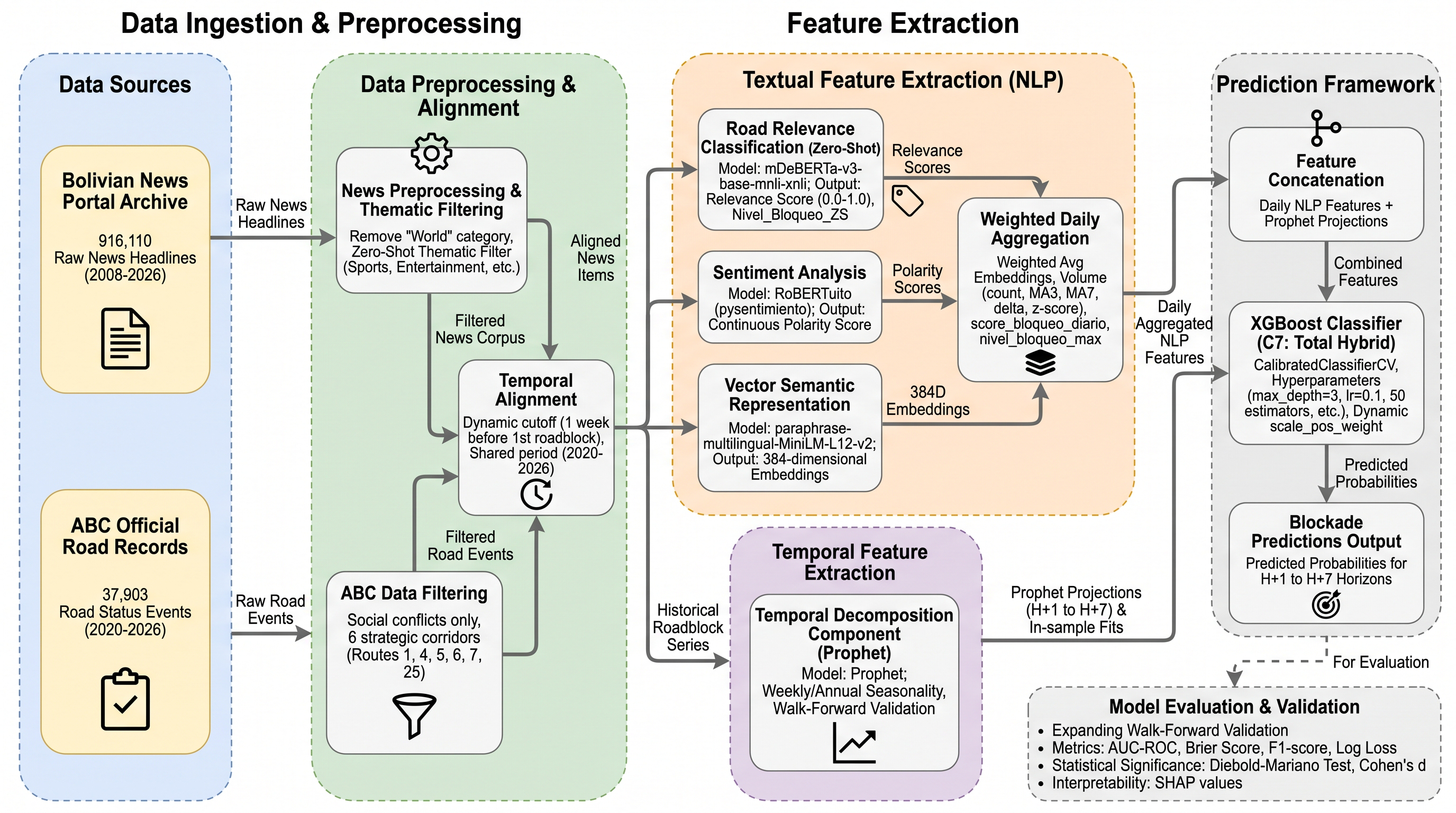}
\caption{Final output generated by the hybrid probabilistic forecasting system.}%
\label{fig:final_output}
\end{figure*}

\subsection{Preprocessing and Thematic Filtering}

The news corpus was cleaned in two phases. First, the ``World'' category was removed due to its lack of relevance to local conflict. Second, we applied a filter based on zero-shot thematic classification (detailed in Section 3.3) to exclude content unrelated to the phenomenon, such as sports, entertainment, horoscope, and fashion. After this process, the resulting corpus was adjusted to 903,602 articles.

To temporally align both sources, a dynamic cutoff criterion was adopted: the start of the time series was set one week before the first roadblock date recorded in the ABC database. This margin ensures that the model has the necessary prior media context without including periods of inactivity that do not provide a predictive signal for the studied phenomenon.

\subsection{Textual Feature Extraction}

Three feature extraction processes were applied to each filtered news item using pre-trained language models:

\noindent\textbf{Road relevance classification (zero-shot).} A multilingual zero-shot classifier (\texttt{mDeBERTa-v3-base-mnli-xnli}) was used to assign each headline a continuous relevance score with respect to four levels of road conflict intensity:

\noindent{}-- Confirmed physical roadblock (reference weight 1.0)\par
\noindent{}-- March or roadblock announcement (0.60)\par
\noindent{}-- Strike or union demand (0.20)\par
\noindent{}-- Unrelated content (0.0).\par
Additionally, the discrete level with the highest probability (\texttt{Nivel\_Bloqueo\_ZS}) was recorded for each news item.

\noindent\textbf{Sentiment analysis.} A Spanish sentiment analysis model (\texttt{RoBERTuito}, via the \texttt{pysentimiento} library) was applied to obtain a continuous polarity score per news item.

\noindent\textbf{Vector semantic representation.} Each headline was transformed into a 384-dimensional vector using the multilingual sentence embeddings model \texttt{paraphrase-multilingual-MiniLM-L12-v2}. A preliminary evaluation was conducted on reducing this dimensionality into a low-dimensional latent space ($d=10$) using Principal Component Analysis (PCA). However, experimental tests demonstrated a systematic degradation in the classifier's discriminative capacity (AUC-ROC), attributed to the massive loss of explained variance inherent in such an aggressive compression of dense semantic embeddings. Consequently, it was decided to preserve the original 384 dimensions to maximize expressive capacity and safeguard the model's decision boundaries.

\subsection{Weighted Daily Aggregation}

Individual news items were aggregated at the daily level using a weighting scheme based on the road relevance score described in 3.3. For each day $d$, the aggregated semantic vector was calculated as the weighted average of the embeddings of all news items published on that day, using the road relevance score as the weight when it was positive, and a differentiated residual weight (0.05 for news explicitly classified as irrelevant, 0.30 for news with no zero-shot classification available) otherwise. This weighting amplifies the contribution of news with content semantically related to an active road conflict compared to unrelated background media coverage.

Additionally, the absolute volume of news, its 3- and 7-day moving averages, its daily variation (delta), and its normalized z-score were calculated as daily variables, along with the daily average of the road relevance score (\texttt{score\_bloqueo\_diario}) and the maximum level of escalation observed in the day (\texttt{nivel\_bloqueo\_max}).

\subsection{Temporal Decomposition Component}

The Prophet model~\cite{22} was employed to capture the seasonal structure of the phenomenon, incorporating weekly and annual seasonality components. To prevent temporal information leakage, a single Prophet model was fitted at each step of the walk-forward validation scheme (Section 3.7) over the history available up to the evaluation day, simultaneously generating the in-sample fitted values (used as a predictor variable in the training set) and the out-of-sample projections for the seven evaluated forecasting horizons.

\subsection{Seven-Level Ablation Architecture}

In order to isolate the incremental contribution of each information source, seven experimental configurations were designed, each trained using a calibrated XGBoost classifier (Platt scaling via \texttt{CalibratedClassifierCV}) with the following uniform hyperparameters: maximum depth of 3, learning rate of 0.1, 50 estimators, column subsampling per tree of 0.3 (implicit regularization given the high dimensionality of the embeddings), row subsampling of 0.8, and a dynamic positive class weight (\texttt{scale\_pos\_weight}) recalculated in each training window as the ratio between negative and positive observations, to compensate for the moderate class imbalance (historical prevalence $\approx$ 38--41\%).

The seven configurations were:

-C1 (Pure Prophet): only the temporal decomposition projection, recalibrated using logistic regression (Platt scaling) on the in-sample fitted values.

-C2 (Volume): only the press volume variables (daily count, moving averages, delta, z-score).

-C3 (Pure NLP): only the aggregated 384-dimensional semantic vector, without the temporal or volume components.

-C4 (Volume + Zero-Shot + NLP): integration of volume, road relevance score, and semantics, without the temporal component.

-C5 (Prophet + Zero-Shot): temporal component combined with volume and road relevance score.

-C6 (Prophet + NLP): temporal component combined with volume and full vector semantics.

-C7 (Total Hybrid): integration of all four information sources (temporal, volume, road relevance score, semantics).

To determine whether the gain observed in configurations C5--C7 is attributable to the proposed information architecture or to the learning algorithm employed, four external benchmarks were additionally evaluated under the same walk-forward scheme and the same horizons:

\noindent{}-- B1 (Regularized Logistic Regression) C=0.01.\par
\noindent{}-- B2 (LightGBM) 100 estimators, depth 3, learning rate 0.05\par
\noindent{}-- B3 (Univariate SARIMA) of order (1,1,1)$\times$ (1,1,1,7), refitted at each step of the walk-forward on the \texttt{hubo\_bloqueo} series\par
\noindent{}-- B4 (XGBoost) with the same features as C7 plus target lags of 1, 2, 3, and 7 days.\par
The hyperparameters of B1, B2, and B4 were selected via grid search over a separate tuning window (August 15, 2021, to December 31, 2023), excluding the final evaluation period (January 1, 2024, to June 11, 2026) to avoid model selection bias. The benchmarks receive the same features as C7, ensuring that any observed difference reflects the algorithm's capability and not information availability.

\subsection{Validation Scheme: Expanding Walk-Forward}

The evaluation was conducted using an expanding window walk-forward validation scheme, considered the methodological standard to prevent information leakage in time series forecasting problems~\cite{23}. For each day $t$ of the evaluation period (August 15, 2021, to June 11, 2026; 1,762 days), the seven configurations were trained using exclusively information available up to day $t-1$, and predictions were generated for the seven horizons $t+1$ to $t+7$. The forecasting target ($Y$) at each horizon $h$ was constructed by shifting the binary roadblock occurrence variable, ensuring that the predictor variable \texttt{inercia\_prophet} used in training corresponded, for each historical observation, to the Prophet projection generated for the impact day corresponding to that horizon, exactly replicating the informational structure available at the time of actual inference.

\subsection{Evaluation Metrics}

Four families of metrics are reported. Discrimination: area under the ROC curve (AUC-ROC). Calibration: Brier Score and Brier Skill Score (BSS) with respect to the historical climatology of the problem, defined as $p(1-p)$ where $p$ is the global prevalence of the positive class. Binary classification: precision, recall, F1-score, and log loss, calculated over the decision threshold that maximizes the F1-score within the out-of-sample set for each horizon. Statistical significance: Diebold–Mariano test (Diebold \& Mariano, 1995) with Newey–West variance correction for multiple horizons, applied to seven paired comparisons of scientific interest, along with the effect size (Cohen's $d$) on the squared error differential between configurations. Additionally, this work includes calibration curves (reliability diagrams) for the established configurations and benchmarks at selected horizons (H+1, H+4, and H+7), providing the opportunity to evaluate the coherence between the emitted probabilities and the corresponding frequencies. The interpretability of the C7 model is also examined using SHAP (SHapley Additive exPlanations) values, calculated on the base XGBoost classifier prior to Platt calibration, with the 384 embedding dimensions aggregated as a single group to facilitate readability.

\section{Results and Discussion}

\subsection{Predictive Performance Evaluation}

The evaluation of the seven experimental configurations under the walk-forward validation scheme evidences a division between two groups of models. Configurations that incorporate the temporal decomposition component (C1, C5, C6, C7) outperform those that do not (C2, C3, C4) in AUC-ROC and Brier Score across nearly all evaluated horizons. Within the first group, the incorporation of semantic news signals on top of the seasonal baseline (C6: Prophet+NLP) produces a consistent improvement over pure Prophet (C1) across all seven horizons, both in discriminative capacity and probabilistic calibration.

\begin{figure*}[htbp]
\centering
\includegraphics[width=\textwidth]{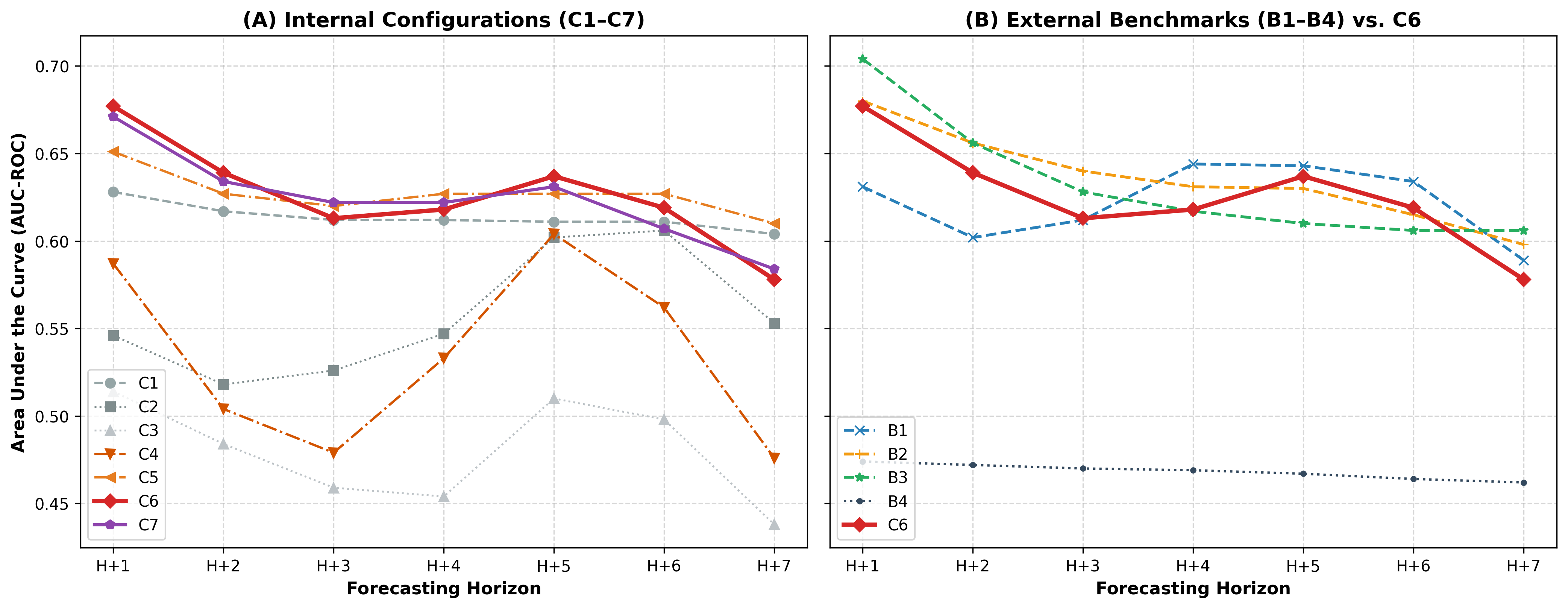}
\caption{Area under the curve (AUC) by forecasting horizon.}%
\label{fig:auc_por_horizonte}
\end{figure*}

The comparison between C6 and C7 (Total Hybrid) reveals no statistically significant difference 
across most horizons. This equivalence occurs because the zero-shot relevance score is already 
implicitly incorporated into C6 as the weighting mechanism during daily embedding aggregation 
(Section 3.4). Consequently, adding the zero-shot score as an explicit tabular feature in C7 
provides redundant information, confirming C6 as the most parsimonious architecture.

The evaluation of the four external benchmarks over the same walk-forward period reveals a distinct pattern that depends on both the horizon and the metric considered. In terms of discrimination (AUC-ROC), the most competitive benchmark in H+1 turns out to be B3 (univariate SARIMA), with AUC=0.704, a value that exceeds C6 (0.677) and C7 (0.671) in that specific horizon. This result reflects the strong short-term autoregressive component in the roadblocks series: when a corridor is disrupted on day t, the probability of disruption on t+1 is substantially higher than the base rate, a signal that SARIMA captures directly without the need for textual information. B2 (LightGBM trained on the same set of features as C7) reaches AUC=0.680 in H+1, which is also higher than or comparable to C6, and maintains competitiveness in H+2 (0.656) and H+3 (0.640).

However, this advantage of the univariate and alternative benchmarks is systematically reversed as the horizon extends. B3 drops from AUC=0.704 in H+1 to 0.606 in H+7, while C6 stands at 0.578, and in Brier Score, C6 consistently outperforms B3 across all horizons. Even more relevant for operational application, B3 presents the highest log loss values in the study (see Table~\ref{tab:tabla_5_resumen_log_loss_general}), showing poor probabilistic calibration: the model emits extreme probabilities that deviate from actual frequencies, directly limiting its utility as a production alert system where the emitted probability is used to scale the logistical response. B4 (XGBoost with autoregressive target lags instead of semantic embeddings) shows the weakest performance among the group of benchmarks, with AUC between 0.462 and 0.474 across all horizons---close to the level of chance---confirming that the temporal inertia of the target alone does not capture the signal that press embeddings provide over medium and long horizons. B1 (regularized Logistic Regression) occupies an intermediate position, with AUC between 0.589 (H+7) and 0.644 (H+4), and log loss comparable to that of C5, suggesting that a linear function over the feature space captures part of the conflict pattern but lacks the ability to model non-linear interactions exploited by C6 and C7.

\begin{figure*}[htbp]
\centering
\includegraphics[width=\textwidth]{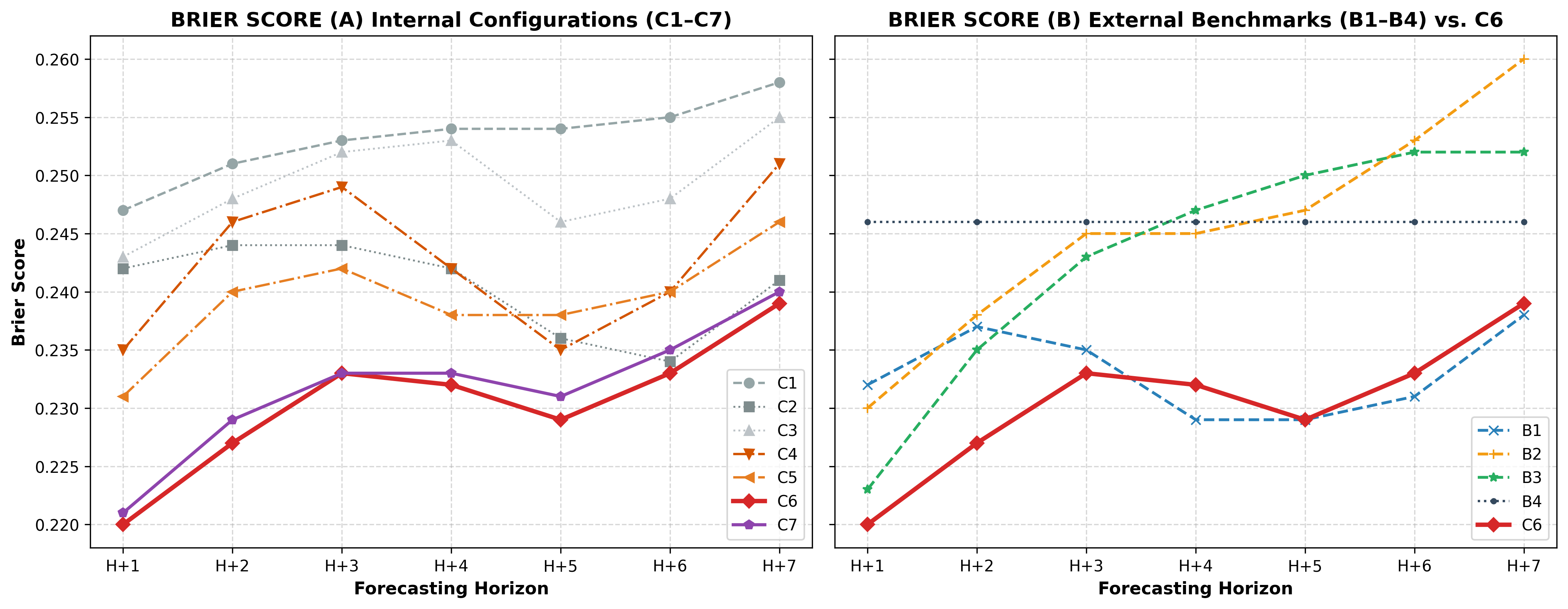}
\caption{Brier Score by forecasting horizon.}%
\label{fig:brier_por_horizonte}
\end{figure*}

\begin{figure*}[htbp]
\centering
\includegraphics[width=\textwidth]{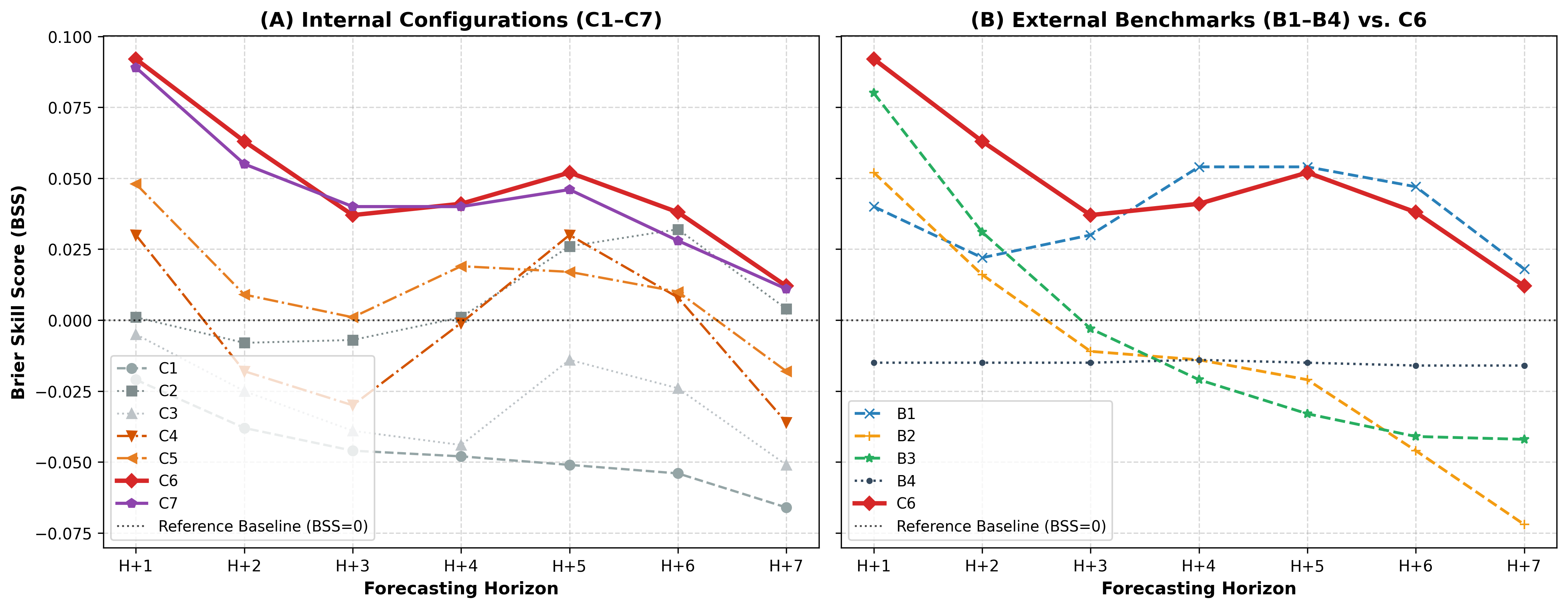}
\caption{Brier Skill Score (BSS) by forecasting horizon.}%
\label{fig:bss_por_horizonte}
\end{figure*}

Hybrid model performance: Configurations C6 and C7 present the lowest Brier Score values (best performance) across most evaluated horizons (H+1 to H+7), indicating higher accuracy in probability estimation compared to the observed reality. In particular, C6 stands out with a positive BSS that is consistently higher than C1 (pure Prophet), validating that the incorporation of semantic press representations adds predictive value by correcting the probabilistic estimations of the seasonal component.

Comparison with Benchmarks: When contrasting with external models, we observe a competitive advantage for C6. Although B3 (SARIMA) shows low Brier Score values in the immediate horizon (H+1: 0.223), its performance degrades rapidly as the horizon expands. This weakness is clearly reflected in the BSS (Table~\ref{tab:tabla_3_bss_score}), where C6 maintains a sustained advantage over B3, which goes from positive values in H+1 to performance deterioration (negative values) starting from H+3. This confirms that the proposed hybrid model offers a more robust and stable probabilistic calibration for medium-term logistical decision-making.

Discriminative capacity and calibration: The consistency of C6 and C7 in maintaining a positive BSS over extended horizons, compared to the volatility of purely autoregressive models (such as B3) or models without a temporal component (C3 and C4), underlines that the combination of seasonality and semantics not only improves discrimination but also optimizes the reliability of the probabilistic alerts generated by the system.

\subsection{Statistical Significance and Robustness of the Diebold–Mariano Test (All Relevant Pairs)}

The Diebold–Mariano test confirms that the incorporation of the vector semantic representation (C6) reduces the forecasting error compared to the pure temporal decomposition model (C1), with p-values below 0.002 across the seven evaluated horizons (Table~\ref{tab:tabla_4_par_c6_vs_c1_nlp_agrega_sobre_pr}). The effect size (Cohen's $d$, between $-0.099$ and $-0.152$) corresponds to a small magnitude according to conventional thresholds, but it is consistent and statistically robust.

\begin{table}[htbp]
\centering
\caption{Pair C6 vs C1 --- Does NLP add value over Prophet?}%
\label{tab:tabla_4_par_c6_vs_c1_nlp_agrega_sobre_pr}
\small
\resizebox{\columnwidth}{!}{%
\begin{tabular}{lcccc}
\toprule
\textbf{H} & \textbf{DM Stat} & \textbf{p-value} & \textbf{Cohen's d} & \textbf{Sig} \\
\midrule
H+1 & -6.3906 & 0 & -0.1522 & ** \\
H+2 & -5.0852 & 0 & -0.1374 & ** \\
H+3 & -3.8954 & 0.0001 & -0.1123 & ** \\
H+4 & -4.0773 & 0 & -0.121 & ** \\
H+5 & -4.505 & 0 & -0.1389 & ** \\
H+6 & -3.8232 & 0.0001 & -0.1224 & ** \\
H+7 & -3.1767 & 0.0015 & -0.0987 & ** \\
\bottomrule
\end{tabular}%
}
\end{table}

\begin{table}[htbp]
\centering
\caption{C6 vs C3 --- Does Prophet+NLP outperform pure NLP?}%
\label{tab:tabla_5_c6_vs_c3_prophet_nlp_supera_nlp_}
\small
\resizebox{\columnwidth}{!}{%
\begin{tabular}{lcccc}
\toprule
\textbf{H} & \textbf{DM Stat} & \textbf{p-value} & \textbf{Cohen's d} & \textbf{Sig} \\
\midrule
H+1 & -6.4075 & 0 & -0.1526 & ** \\
H+2 & -4.8619 & 0 & -0.1316 & ** \\
H+3 & -4.2366 & 0 & -0.1232 & ** \\
H+4 & -4.295 & 0 & -0.1287 & ** \\
H+5 & -4.5509 & 0 & -0.1409 & ** \\
H+6 & -3.5475 & 0.0004 & -0.1151 & ** \\
H+7 & -3.3702 & 0.0008 & -0.1054 & ** \\
\bottomrule
\end{tabular}%
}
\end{table}

The contrast between C6 and C3 (pure NLP, without a temporal component) is informative, revealing the largest effect size in the entire study (Cohen's $d$ between -0.147 and -0.201, $p < 0.0001$ across all horizons). This demonstrates that the semantic press representation requires the anchoring provided by the seasonal component to constitute a useful predictive signal; on its own, this representation does not reliably discriminate between days with and without roadblocks, as confirmed by its AUC being close to or below the level of chance (0.44--0.51) across most horizons.

\begin{table}[htbp]
\centering
\caption{C7 vs C6 --- Does ZS add value over Prophet+NLP\@? [ZS RELEVANCE]}%
\label{tab:tabla_6_c7_vs_c6_zs_agrega_sobre_prophet}
\small
\resizebox{\columnwidth}{!}{%
\begin{tabular}{lcccc}
\toprule
\textbf{H} & \textbf{DM Stat} & \textbf{p-value} & \textbf{Cohen's d} & \textbf{Sig} \\
\midrule
H+1 & 0.5015 & 0.616 & 0.0119 &  \\
H+2 & 1.6527 & 0.0984 & 0.0413 &  \\
H+3 & -0.5388 & 0.59 & -0.0129 &  \\
H+4 & 0.3105 & 0.7562 & 0.0075 &  \\
H+5 & 1.302 & 0.1929 & 0.0344 &  \\
H+6 & 2.3191 & 0.0204 & 0.0617 & * \\
H+7 & 0.2906 & 0.7713 & 0.0073 &  \\
\bottomrule
\end{tabular}%
}
\end{table}

\begin{table}[htbp]
\centering
\caption{C7 vs B1 --- Does the hybrid outperform Logistic Regression? [LINEAR BENCHMARK]}%
\label{tab:tabla_7_c7_vs_b1_h_brido_supera_a_logist}
\small
\resizebox{\columnwidth}{!}{%
\begin{tabular}{lcccc}
\toprule
\textbf{H} & \textbf{DM Stat} & \textbf{p-value} & \textbf{Cohen's d} & \textbf{Sig} \\
\midrule
H+1 & -3.1061 & 0.0019 & -0.074 & ** \\
H+2 & -2.1917 & 0.0284 & -0.0608 & * \\
H+3 & -0.6046 & 0.5454 & -0.0168 &  \\
H+4 & 0.8404 & 0.4007 & 0.024 &  \\
H+5 & 0.4642 & 0.6425 & 0.0138 &  \\
H+6 & 1.3086 & 0.1907 & 0.0378 &  \\
H+7 & 0.5726 & 0.5669 & 0.0166 &  \\
\bottomrule
\end{tabular}%
}
\end{table}

\begin{table}[htbp]
\centering
\caption{C7 vs B2 --- Does the hybrid outperform the equivalent LightGBM\@? [ALT\@. GBM BENCHMARK]}%
\label{tab:tabla_8_c7_vs_b2_h_brido_supera_a_lightg}
\small
\resizebox{\columnwidth}{!}{%
\begin{tabular}{lcccc}
\toprule
\textbf{H} & \textbf{DM Stat} & \textbf{p-value} & \textbf{Cohen's d} & \textbf{Sig} \\
\midrule
H+1 & -2.6923 & 0.0071 & -0.0641 & ** \\
H+2 & -2.4959 & 0.0126 & -0.0651 & * \\
H+3 & -2.8868 & 0.0039 & -0.0806 & ** \\
H+4 & -3.0339 & 0.0024 & -0.0861 & ** \\
H+5 & -3.3887 & 0.0007 & -0.1056 & ** \\
H+6 & -3.5638 & 0.0004 & -0.1095 & ** \\
H+7 & -3.8739 & 0.0001 & -0.1189 & ** \\
\bottomrule
\end{tabular}%
}
\end{table}

\begin{table}[htbp]
\centering
\caption{C7 vs B3 --- Does the hybrid outperform univariate SARIMA\@? [ARIMA BENCHMARK]}%
\label{tab:tabla_9_c7_vs_b3_h_brido_supera_a_sarima}
\small
\resizebox{\columnwidth}{!}{%
\begin{tabular}{lcccc}
\toprule
\textbf{H} & \textbf{DM Stat} & \textbf{p-value} & \textbf{Cohen's d} & \textbf{Sig} \\
\midrule
H+1 & -0.4433 & 0.6575 & -0.0106 &  \\
H+2 & -1.3592 & 0.1741 & -0.035 &  \\
H+3 & -2.1855 & 0.0289 & -0.0608 & * \\
H+4 & -2.99 & 0.0028 & -0.0864 & ** \\
H+5 & -3.8499 & 0.0001 & -0.1154 & ** \\
H+6 & -3.0616 & 0.0022 & -0.0958 & ** \\
H+7 & -2.2799 & 0.0226 & -0.0719 & * \\
\bottomrule
\end{tabular}%
}
\end{table}

\begin{table}[htbp]
\centering
\caption{C7 vs B4 --- Does NLP add value over Lag Features+XGBoost? [AUTOREGRESSIVE BENCHMARK]}%
\label{tab:tabla_10_c7_vs_b4_nlp_agrega_sobre_lag_f}
\small
\resizebox{\columnwidth}{!}{%
\begin{tabular}{lcccc}
\toprule
\textbf{H} & \textbf{DM Stat} & \textbf{p-value} & \textbf{Cohen's d} & \textbf{Sig} \\
\midrule
H+1 & -7.5015 & 0 & -0.1787 & ** \\
H+2 & -4.6182 & 0 & -0.1287 & ** \\
H+3 & -3.4137 & 0.0006 & -0.1017 & ** \\
H+4 & -3.3957 & 0.0007 & -0.1026 & ** \\
H+5 & -3.7649 & 0.0002 & -0.1182 & ** \\
H+6 & -3.0086 & 0.0026 & -0.0951 & ** \\
H+7 & -1.8815 & 0.0599 & -0.0616 &  \\
\bottomrule
\end{tabular}%
}
\end{table}

The Diebold–Mariano tests against the four external benchmarks (Tables~\ref{tab:tabla_7_c7_vs_b1_h_brido_supera_a_logist}--\ref{tab:tabla_10_c7_vs_b4_nlp_agrega_sobre_lag_f}) directly address whether the superiority of C7 over C1 is attributable to the proposed information architecture or to the learning algorithm employed. Against B2 (LightGBM trained on the same set of features as C7), the DM test yields significant values across all seven horizons ($p < 0.05$ in all cases, $p < 0.01$ in most), ruling out that the observed gain of C7 over C1 can be explained by an inherent advantage of the XGBoost algorithm over other gradient boosting methods: when the information set is identical, XGBoost calibrated with temporal decomposition consistently outperforms LightGBM\@. The comparison against B4 (XGBoost with autoregressive target lags instead of semantic embeddings) is also significant from H+1 to H+6, with Cohen's $d$ between $-0.08$ and $-0.15$, confirming that the semantic signal extracted from press embeddings provides information not captured by the mere temporal inertia of the target expressed as direct lags.

Against B1 (Logistic Regression) and B3 (SARIMA), the pattern is more nuanced. C7 significantly outperforms B1 in H+1 and H+2 ($p < 0.05$), but the difference does not reach statistical significance in H+3 to H+7, which is consistent with the competitiveness of B1 in F1-score over medium horizons observed in the experiments. Against B3 (SARIMA), C7 shows a significant advantage in H+3 to H+7 but not in H+1, where the immediate autoregressive component favors the univariate model—a finding consistent with the higher AUC of B3 in that specific horizon already discussed in Section 4.1. Taken together, these results suggest that the contribution of the semantic component is more robust against gradient boosting alternatives than against purely temporal models in short horizons, and that the advantage of the proposed hybrid system consolidates from H+3 onward, where the seasonal structure of Prophet combined with the semantic signal from the press captures information that univariate benchmarks cannot represent.

\begin{table}[htbp]
\centering
\caption{Longest streak without roadblocks: 41 days. Period: November 15 to December 25, 2024 (objective criterion: maximum consecutive streak free in the historical data)}%
\label{tab:tabla_11_racha_m_s_larga_sin_bloqueos_41}
\small
\resizebox{\columnwidth}{!}{%
\begin{tabular}{lcccc}
\toprule
\textbf{H} & \textbf{Config} & \textbf{Mean Prob (Free)} & \textbf{Mean Prob (Block)} & \textbf{Separation} \\
\midrule
H+1 & C1 & 0.3957 & 0.4988 & 0.1031 \\
 & C2 & 0.3786 & 0.3916 & 0.013 \\
 & C3 & 0.3964 & 0.4029 & 0.0065 \\
 & C4 & 0.3937 & 0.4211 & 0.0274 \\
 & C5 & 0.4007 & 0.4987 & 0.098 \\
 & C6 & 0.3953 & 0.483 & 0.0877 \\
 & C7 & 0.3949 & 0.4867 & 0.0918 \\
 & B1 & 0.3541 & 0.4151 & 0.061 \\
 & B2 & 0.3831 & 0.5405 & 0.1574 \\
 & B3 & 0.3363 & 0.5346 & 0.1983 \\
 & B4 & 0.3864 & 0.3838 & -0.0026 \\
H+3 & C1 & 0.4024 & 0.4931 & 0.0907 \\
 & C2 & 0.3805 & 0.3864 & 0.0059 \\
 & C3 & 0.3959 & 0.386 & -0.0099 \\
 & C4 & 0.3994 & 0.3935 & -0.0059 \\
 & C5 & 0.4089 & 0.4899 & 0.0809 \\
 & C6 & 0.3995 & 0.4497 & 0.0502 \\
 & C7 & 0.4033 & 0.4626 & 0.0593 \\
 & B1 & 0.3651 & 0.4131 & 0.048 \\
 & B2 & 0.3926 & 0.5102 & 0.1176 \\
 & B3 & 0.3805 & 0.4785 & 0.098 \\
 & B4 & 0.3872 & 0.3845 & -0.0027 \\
H+7 & C1 & 0.4067 & 0.4934 & 0.0866 \\
 & C2 & 0.378 & 0.3931 & 0.0151 \\
 & C3 & 0.3861 & 0.3719 & -0.0142 \\
 & C4 & 0.3852 & 0.378 & -0.0072 \\
 & C5 & 0.4103 & 0.4874 & 0.0772 \\
 & C6 & 0.3908 & 0.4199 & 0.0291 \\
 & C7 & 0.3956 & 0.4304 & 0.0348 \\
 & B1 & 0.3774 & 0.4126 & 0.0352 \\
 & B2 & 0.4032 & 0.4856 & 0.0824 \\
 & B3 & 0.3892 & 0.4668 & 0.0776 \\
 & B4 & 0.388 & 0.3845 & -0.0035 \\
\bottomrule
\end{tabular}%
}
\end{table}

As a sanity check to verify logical discriminative consistency, the system's performance was evaluated using the class probability separation margin, defined as $\Delta P = P(\text{Blocked}) - P(\text{Free})$. A valid probabilistic forecasting model must satisfy $\Delta P > 0$, ensuring it assigns systematically higher risk probabilities to actual roadblock events than to free days.

The system's behavior during prolonged periods of social calm was evaluated over the longest streak of consecutive days without roadblocks recorded in the historical data (41 days, spanning November and December 2024). Within this specific test window, temporally anchored configurations ($C1$, $C5$, $C6$, $C7$) consistently maintain positive separation across all evaluated horizons ($H+1$, $H+3$, $H+7$). This confirms that their discriminative capacity does not rely exclusively on the historical prevalence of the positive class. In particular, the proposed architecture ($C6$) maintains stable positive margins ($+0.088$ at $H+1$, $+0.050$ at $H+3$, and $+0.029$ at $H+7$), proving that its signal stems from genuine pre-conflict dynamics rather than false alarms during calm periods.

In contrast, naive NLP-only and improperly integrated hybrid setups ($C2\text{--}C4$), along with the persistent baseline ($B4$), fail this sanity check. Configurations $C3$ and $C4$ exhibit negative separation at $H+3$ and $H+7$ (reaching $-0.014$ for $C3$ at $H+7$), erroneously assigning on average a higher probability to effectively free days than to days with roadblocks within this window—further evidence of their limited discriminative capacity in the absence of the temporal component. Furthermore, while tree-based benchmarks ($B2$, $B3$) display larger raw separation at $H+1$ due to extreme probability estimates near $0$ and $1$, this overconfidence leads to severe calibration degradation over extended horizons. Configuration $C6$ strikes the optimal balance between robust class separation and long-term probabilistic calibration.

\begin{figure*}[htbp]
\centering
\includegraphics[width=\textwidth]{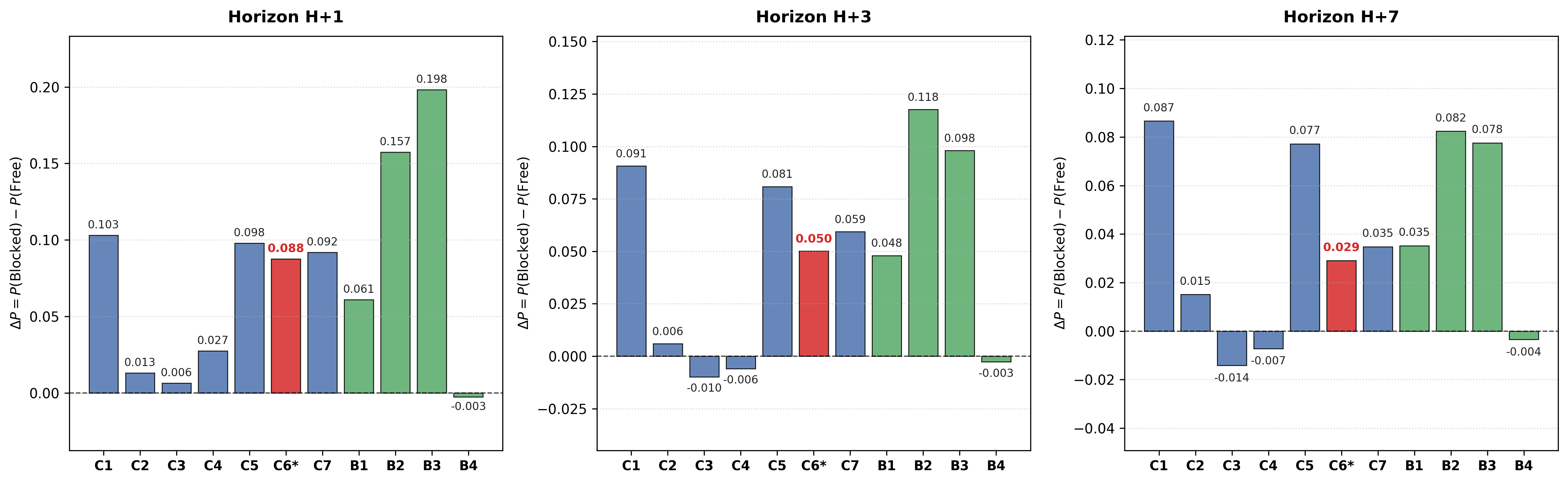}
\caption{Separation capacity of the distinct configurations between roadblock days and free days.}%
\label{fig:sanity_check_separacion}
\end{figure*}

\subsection{Classification and Calibration Metrics}

The analysis of the confusion matrix and derived classification metrics (Section 6D) confirms the same pattern observed in the probabilistic metrics: configurations C3, C4, and, to a lesser extent, C2 tend to collapse toward an almost uniform prediction of the positive class across multiple horizons (recall close to 1.0 with specificity close to 0), a behavior consistent with the absence of a real discriminative signal rather than effective classifier performance. In contrast, configurations C5, C6, and C7 exhibit confusion matrices with substantial representation across all four classification categories, achieving the highest F1-score values in the study (C6: 0.583–0.617) under the optimal decision threshold determined on the out-of-sample set for each horizon.

\subsection{Operational Risk Radar: Application Case Study}

As an operational validation exercise, the system was run on the week of June 12 to 18, 2026, generating out-of-sample forecasts for the seven evaluated horizons. Configuration C6 projected roadblock probabilities substantially higher than the base rate of the last 365 days (30.7\%) across the entirety of the analyzed horizon, reaching ``High Risk'' according to the 80th percentile threshold of the historical distribution of predictions on all days except H+7. Noteworthy is the divergence between the C6 projection and that of the isolated calibrated Prophet component (C1), which in horizons H+2 and H+3 projected probabilities below the base rate (18.97\% and 14.54\%, respectively) in contrast to the probabilities above 70\% emitted by C6. This divergence illustrates the mechanism proposed in this work: the incorporation of semantic press signals allows the system to detect escalations in social tension not captured by the pure seasonal structure, whose behavior is rigorously validated by monitoring the actual materialization of these forecasts in subsequent days.

\begin{table*}[htbp]
\centering
\setlength{\fboxsep}{6pt}
\fbox{%
\begin{minipage}{\dimexpr\textwidth-2\fboxsep-2\fboxrule\relax}
\centering

\begin{minipage}{0.49\textwidth}
\centering
\caption{Weekly Roadblock Risk}%
\label{tab:tabla_12_riesgo_de_bloqueo_semanal}
\scriptsize
\setlength{\tabcolsep}{3pt}
\resizebox{\linewidth}{!}{%
\begin{tabular}{lccccccc}
\toprule
\textbf{H} & \textbf{Date} & \textbf{Prob C6} & \textbf{Prob C7} & \textbf{C1 Calib.} & \textbf{Mod Thresh} & \textbf{P80 Thresh} & \textbf{Status (C7)} \\
\midrule
H+1 & Jun.~12 & 0.76 & 0.75 & 0.34 & 0.31 & 0.57 & HIGH RISK (P80: 0.56) \\
H+2 & Jun.~13 & 0.74 & 0.63 & 0.19 & 0.31 & 0.55 & HIGH RISK (P80: 0.55) \\
H+3 & Jun.~14 & 0.68 & 0.64 & 0.15 & 0.31 & 0.56 & HIGH RISK (P80: 0.55\%) \\
H+4 & Jun.~15 & 0.65 & 0.64 & 0.51 & 0.31 & 0.55 & HIGH RISK (P80: 0.54) \\
H+5 & Jun.~16 & 0.65 & 0.61 & 0.47 & 0.31 & 0.54 & HIGH RISK (P80: 0.53) \\
H+6 & Jun.~17 & 0.64 & 0.61 & 0.43 & 0.31 & 0.52 & HIGH RISK (P80: 0.52) \\
H+7 & Jun.~18 & 0.58 & 0.51 & 0.46 & 0.31 & 0.52 & MODERATE RISK (Base: 0.30) \\
\bottomrule
\end{tabular}%
}
\vspace{0.3em}
\raggedright\scriptsize HIGH RISK = High risk; MODERATE RISK = Moderate risk.
\end{minipage}
\hfill
\begin{minipage}{0.48\textwidth}
\centering
\includegraphics[width=\textwidth]{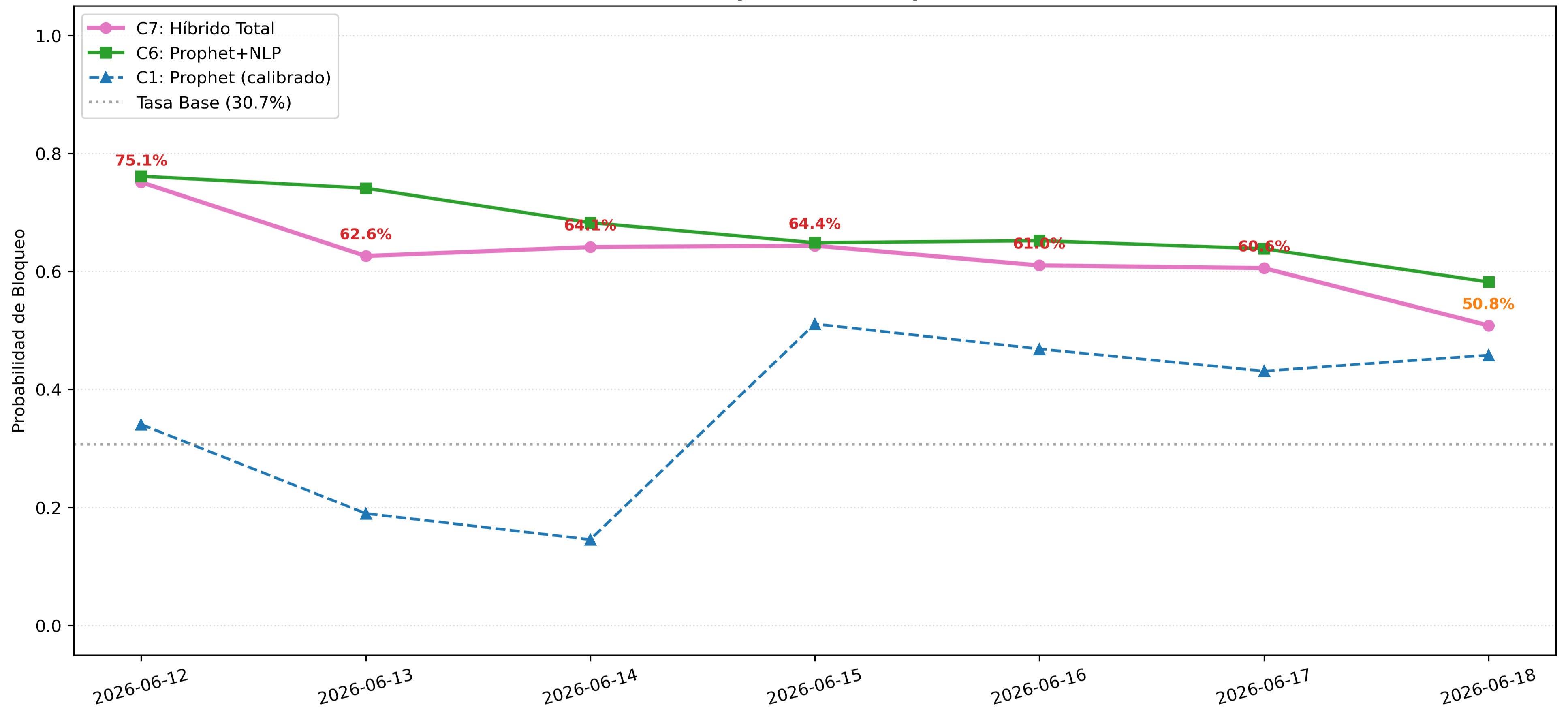}
\captionof{figure}{Weekly operational risk on critical routes according to configurations C6 and C7.}%
\label{fig:radar_semanal_c6_c7}
\end{minipage}

\end{minipage}%
}
\end{table*}

Although B3 (SARIMA) and B2 (LightGBM) exhibit competitive or superior AUC compared to C6 in H+1, their selection as an operational system is discouraged for two reasons: B3 presents the poorest probabilistic calibration in the study (log loss 0.80–0.86), which limits the interpretability of its probabilities as actionable risk measures; and both models show consistent degradation from horizon H+3 onward, while C6 maintains a lower Brier Score across all seven horizons. The operational radar presented in Figure 4 utilizes C6 as the primary configuration precisely due to this combination of probabilistic calibration and multi-horizon consistency.

\begin{figure*}[htbp]
\centering
\includegraphics[width=\textwidth]{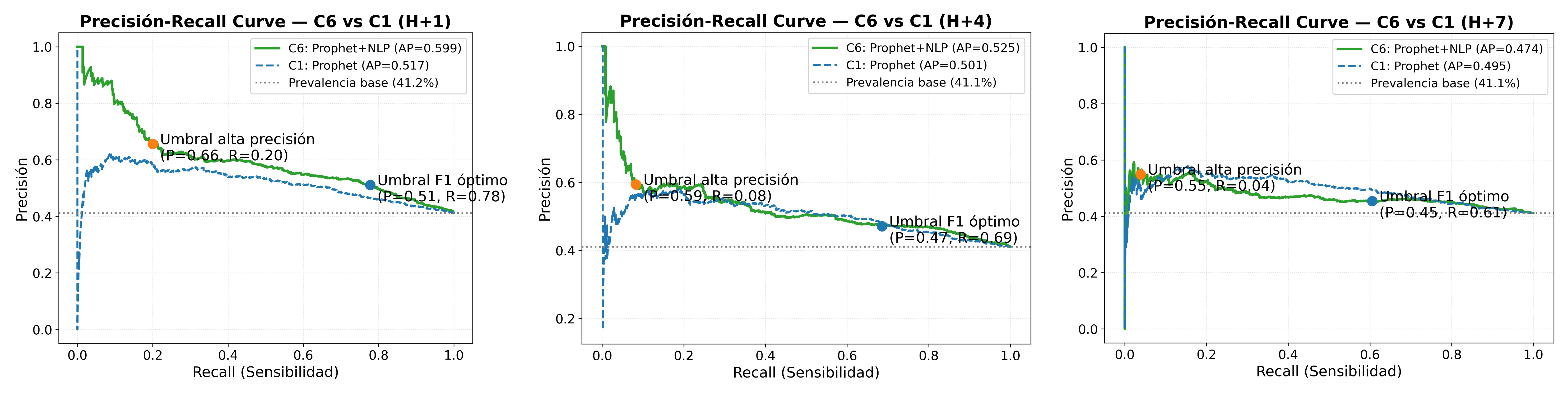} 
\caption{Precision-Recall curves C6 vs C1 (H+1, H+4, H+7).}%
\label{fig:curva_pr_combined}
\end{figure*}

A key aspect for the operational implementation of the system is decision threshold management. The Precision-Recall curves (Figure~\ref{fig:curva_pr_combined}) illustrate the trade-off available in horizons H+1, H+4, and H+7. In H+1, C6 (Average Precision=0.599) consistently outperforms C1 (AP=0.517) across the entire Precision-Recall space: under the threshold optimized for F1 (0.39), C6 detects 78\% of actual roadblocks with a precision of 51\%, whereas raising the threshold to 0.60 yields a precision of 66\% at the cost of reducing recall to 20\%—operationally useful for high fixed-cost decisions such as heavy cargo rerouting. In H+4, the advantage of C6 (AP=0.525) over C1 (AP=0.501) is maintained, though reduced, with both curves converging in the medium recall range. In H+7, C1 (AP=0.495) recovers a slight advantage over C6 (AP=0.474) in binary discrimination, a result consistent with Prophet's competitiveness in AUC at that horizon—although C6 retains its superiority in Brier Score, confirming that the advantage of the semantic component shifts from discrimination toward probabilistic calibration as the horizon extends. The calibration of C6 is precisely what enables this threshold adjustment: a poorly calibrated model cannot offer such a decision surface, regardless of its AUC\@.

\section{Discussion and Limitations}

This work demonstrates that the integration of semantic news analysis provides a statistically significant improvement over models configured purely with temporal components (C6 vs. C1). Roadblock conflicts in Bolivia exhibit a strong seasonal structure captured by baseline time-series models; however, the addition of dense vector embeddings in C6 smooths out false seasonal peaks while providing an overall structural balance to the forecasts. The superior performance of C6 relative to the remaining internal configurations (C1–C5) and external baselines (B1–B4) confirms that its predictive value stems from the complementary interplay between temporal inertia and semantic signals. Notably, comparing configuration C6 to C7 (which explicitly incorporates zero-shot classification scores) reveals that adding explicit zero-shot metrics introduces redundant information. The dense vector embeddings already capture sufficient semantic relevance without requiring additional explicit categorization, validating C6 as the most parsimonious and effective hybrid architecture.

Despite these positive empirical findings, several methodological, temporal, and spatial limitations must be acknowledged:

Spatial Resolution and Corridor Granularity: While the model offers robust risk estimates for the primary Bolivian road network—specifically covering critical transport corridors such as Routes 1,4, 5, 6, 7, and 25—it currently treats all critical routes as a single aggregated target. This macro-level approach introduces a spatial dilution bias, as conflict dynamics in regions like the Chapare differ fundamentally from those in Santa Cruz, Potosí, or La Paz. Disaggregating these critical routes into specific geographic segments will be essential in future iterations to deliver more granular and location-specific alerts.

Media Aggregator Bias: The textual dataset relies exclusively on the bolivian news aggregator portal. Consequently, the system operates under the potential partiality and editorial framing inherent to this specific outlet. The model effectively captures social conflict as filtered and reported by this portal rather than representing an absolute ground truth of social unrest. Benchmarking against multi-source media feeds will be required to quantify and mitigate single-aggregator bias.

Reporting Latency and Real-Time Agility: Although the expanding walk-forward validation scheme strictly prevents data leakage ($X_t \to Y_{t+h}$), the inherent latency of journalistic data can constrain immediate response agility. News articles published on day $t$ may describe sudden, unannounced emergency blockades that materialized earlier the same morning. While the system excels at predicting multi-day conflict escalation horizons, intraday reporting delays limit its utility for real-time emergency rerouting during spontaneous events.

Domain Adaptation, Local Idioms, and Political Regime Shifts: The processing of embeddings relies on general pre-trained language models that may not fully capture local idiomatic expressions, colloquialisms, or jargon specific to the Bolivian press. Furthermore, media discourse undergoes shifting semantics during political transitions, such as the establishment of the new government at the end of 2025. Implementing periodic fine-tuning and continuous domain adaptation represents a critical avenue to maintain a reliable semantic signal across evolving political landscapes.

Socio-Political and Ethical Considerations: Roadblocks in Bolivia are complex human phenomena rooted in power dynamics, collective identity, and structural demands, often preceded by unrecorded local assemblies or union negotiations. Machine learning models capture discursive escalation in media but cannot comprehend the underlying socio-political causes. Crucially, this predictive system is designed strictly as a logistical decision-support tool for supply chain resilience and transport safety; it should not be used to criminalize social protest or substitute for constructive political dialogue.

\section{Conclusions}

This research validated the viability and superiority of a hybrid framework for forecasting and early detection of roadblocks on critical transport routes in Bolivia. The statistical robustness of temporal models was integrated with the necessary context derived from news headlines. Through a walk-forward framework spanning approximately 1,700 days, it was demonstrated that incorporating vector semantic representation and analysis (C6) consistently outperforms univariate statistical models or pure machine learning approaches.

The fundamental contributions of this work are threefold:

It was verified that the predictive value does not reside solely in a seasonal configuration; rather, the semantic and hybrid contribution sustains itself over medium-term horizons, unlike autoregressive models.

Vector representations efficiently capture the intensity and nuances of conflict without the need for explicit categorization. This optimizes the operational efficiency of the system by reducing additional processing.

\onecolumn
\appendix
\section*{Supplementary Material}

\begin{table*}[htbp]
\centering
\caption{Area Under the Curve (AUC)}%
\label{tab:tabla_1_area_under_the_curve_auc}
\footnotesize
\setlength{\tabcolsep}{4pt}
\resizebox{\textwidth}{!}{%
\begin{tabular}{lccccccc}
\toprule
\textbf{Config./Horizon} & \textbf{H+1} & \textbf{H+2} & \textbf{H+3} & \textbf{H+4} & \textbf{H+5} & \textbf{H+6} & \textbf{H+7} \\
\midrule
C1 & 0.628 & 0.617 & 0.612 & 0.612 & 0.611 & 0.611 & 0.604 \\
C2 & 0.546 & 0.518 & 0.526 & 0.547 & 0.602 & 0.606 & 0.553 \\
C3 & 0.514 & 0.484 & 0.459 & 0.454 & 0.510 & 0.498 & 0.438 \\
C4 & 0.587 & 0.504 & 0.479 & 0.533 & 0.604 & 0.562 & 0.476 \\
C5 & 0.651 & 0.627 & 0.620 & 0.627 & 0.627 & 0.627 & 0.610 \\
C6 & 0.677 & 0.639 & 0.613 & 0.618 & 0.637 & 0.619 & 0.578 \\
C7 & 0.671 & 0.634 & 0.622 & 0.622 & 0.631 & 0.607 & 0.584 \\
B1 & 0.631 & 0.602 & 0.612 & 0.644 & 0.643 & 0.634 & 0.589 \\
B2 & 0.680 & 0.656 & 0.640 & 0.631 & 0.630 & 0.615 & 0.598 \\
B3 & 0.704 & 0.656 & 0.628 & 0.617 & 0.610 & 0.606 & 0.606 \\
B4 & 0.474 & 0.472 & 0.470 & 0.469 & 0.467 & 0.464 & 0.462 \\
\bottomrule
\end{tabular}%
}
\end{table*}

\begin{table*}[htbp]
\centering
\caption{Brier Score}%
\label{tab:tabla_2_brier_score}
\footnotesize
\setlength{\tabcolsep}{4pt}
\resizebox{\textwidth}{!}{%
\begin{tabular}{lccccccc}
\toprule
\textbf{Config./Horizon} & \textbf{H+1} & \textbf{H+2} & \textbf{H+3} & \textbf{H+4} & \textbf{H+5} & \textbf{H+6} & \textbf{H+7} \\
\midrule
C1 & 0.247 & 0.251 & 0.253 & 0.254 & 0.254 & 0.255 & 0.258 \\
C2 & 0.242 & 0.244 & 0.244 & 0.242 & 0.236 & 0.234 & 0.241 \\
C3 & 0.243 & 0.248 & 0.252 & 0.253 & 0.246 & 0.248 & 0.255 \\
C4 & 0.235 & 0.246 & 0.249 & 0.242 & 0.235 & 0.240 & 0.251 \\
C5 & 0.231 & 0.240 & 0.242 & 0.238 & 0.238 & 0.240 & 0.246 \\
C6 & 0.220 & 0.227 & 0.233 & 0.232 & 0.229 & 0.233 & 0.239 \\
C7 & 0.221 & 0.229 & 0.233 & 0.233 & 0.231 & 0.235 & 0.240 \\
B1 & 0.232 & 0.237 & 0.235 & 0.229 & 0.229 & 0.231 & 0.238 \\
B2 & 0.230 & 0.238 & 0.245 & 0.245 & 0.247 & 0.253 & 0.260 \\
B3 & 0.223 & 0.235 & 0.243 & 0.247 & 0.250 & 0.252 & 0.252 \\
B4 & 0.246 & 0.246 & 0.246 & 0.246 & 0.246 & 0.246 & 0.246 \\
\bottomrule
\end{tabular}%
}
\end{table*}

\begin{table*}[htbp]
\centering
\caption{BSS Score}%
\label{tab:tabla_3_bss_score}
\scriptsize
\setlength{\tabcolsep}{2pt}
\resizebox{\textwidth}{!}{%
\begin{tabular}{lccccccc}
\toprule
\textbf{Config./Horizon} & \textbf{H+1} & \textbf{H+2} & \textbf{H+3} & \textbf{H+4} & \textbf{H+5} & \textbf{H+6} & \textbf{H+7} \\
\midrule
C1 & -0.021 & -0.038 & -0.046 & -0.048 & -0.051 & -0.054 & -0.066 \\
C2 & 0.001 & -0.008 & -0.007 & 0.001 & 0.026 & 0.032 & 0.004 \\
C3 & -0.005 & -0.025 & -0.039 & -0.044 & -0.014 & -0.024 & -0.051 \\
C4 & 0.030 & -0.018 & -0.030 & -0.001 & 0.030 & 0.008 & -0.036 \\
C5 & 0.048 & 0.009 & 0.001 & 0.019 & 0.017 & 0.010 & -0.018 \\
C6 & 0.092 & 0.063 & 0.037 & 0.041 & 0.052 & 0.038 & 0.012 \\
C7 & 0.089 & 0.055 & 0.040 & 0.040 & 0.046 & 0.028 & 0.011 \\
B1 & 0.040 & 0.022 & 0.030 & 0.054 & 0.054 & 0.047 & 0.018 \\
B2 & 0.052 & 0.016 & -0.011 & -0.014 & -0.021 & -0.046 & -0.072 \\
B3 & 0.080 & 0.031 & -0.003 & -0.021 & -0.033 & -0.041 & -0.042 \\
B4 & -0.015 & -0.015 & -0.015 & -0.014 & -0.015 & -0.016 & -0.016 \\
\bottomrule
\end{tabular}%
}
\end{table*}

\begin{table*}[htbp]
\centering
\caption{F1-score Summary (Overall)}%
\label{tab:tabla_4_resumen_f1_score_general}
\scriptsize
\resizebox{\textwidth}{!}{%
\begin{tabular}{lccccccc}
\toprule
\textbf{Config./Horizon} & \textbf{H+1} & \textbf{H+2} & \textbf{H+3} & \textbf{H+4} & \textbf{H+5} & \textbf{H+6} & \textbf{H+7} \\
\midrule
C1 & 0.591 & 0.586 & 0.585 & 0.584 & 0.583 & 0.585 & 0.583 \\
C2 & 0.590 & 0.584 & 0.583 & 0.585 & 0.593 & 0.599 & 0.589 \\
C3 & 0.584 & 0.582 & 0.583 & 0.583 & 0.581 & 0.583 & 0.584 \\
C4 & 0.583 & 0.583 & 0.582 & 0.587 & 0.585 & 0.583 & 0.582 \\
C5 & 0.605 & 0.597 & 0.595 & 0.597 & 0.596 & 0.602 & 0.595 \\
C6 & 0.617 & 0.601 & 0.591 & 0.596 & 0.605 & 0.600 & 0.589 \\
C7 & 0.614 & 0.606 & 0.598 & 0.594 & 0.601 & 0.601 & 0.584 \\
B1 & 0.606 & 0.584 & 0.586 & 0.604 & 0.606 & 0.606 & 0.592 \\
B2 & 0.617 & 0.604 & 0.602 & 0.593 & 0.597 & 0.591 & 0.586 \\
B3 & 0.638 & 0.595 & 0.581 & 0.579 & 0.577 & 0.579 & 0.578 \\
B4 & 0.581 & 0.582 & 0.582 & 0.583 & 0.583 & 0.583 & 0.583 \\
\bottomrule
\end{tabular}%
}
\end{table*}

\begin{table*}[htbp]
\centering
\caption{Log Loss Summary (Overall)}%
\label{tab:tabla_5_resumen_log_loss_general}
\scriptsize
\resizebox{\textwidth}{!}{%
\begin{tabular}{lccccccc}
\toprule
\textbf{Config./Horizon} & \textbf{H+1} & \textbf{H+2} & \textbf{H+3} & \textbf{H+4} & \textbf{H+5} & \textbf{H+6} & \textbf{H+7} \\
\midrule
C1 & 0.702 & 0.712 & 0.717 & 0.719 & 0.721 & 0.723 & 0.731 \\
C2 & 0.677 & 0.682 & 0.681 & 0.677 & 0.664 & 0.660 & 0.675 \\
C3 & 0.681 & 0.691 & 0.698 & 0.701 & 0.685 & 0.691 & 0.705 \\
C4 & 0.663 & 0.688 & 0.693 & 0.678 & 0.662 & 0.674 & 0.697 \\
C5 & 0.653 & 0.677 & 0.681 & 0.670 & 0.671 & 0.674 & 0.690 \\
C6 & 0.630 & 0.645 & 0.658 & 0.656 & 0.651 & 0.657 & 0.672 \\
C7 & 0.631 & 0.649 & 0.657 & 0.658 & 0.654 & 0.663 & 0.673 \\
B1 & 0.658 & 0.667 & 0.664 & 0.652 & 0.651 & 0.654 & 0.668 \\
B2 & 0.663 & 0.684 & 0.704 & 0.705 & 0.711 & 0.725 & 0.741 \\
B3 & 0.820 & 0.801 & 0.799 & 0.815 & 0.830 & 0.854 & 0.856 \\
B4 & 0.685 & 0.685 & 0.685 & 0.685 & 0.685 & 0.686 & 0.686 \\
\bottomrule
\end{tabular}%
}
\end{table*}

\clearpage
\subsection*{Confusion Matrix and Classification Metrics --- C1–C7}

Optimal threshold: maximizes F1 on the OOS set for horizon H+1

\begin{table*}[htbp]
\centering
\caption{Table H+1}%
\label{tab:tabla_h1}
\scriptsize
\resizebox{\textwidth}{!}{%
\begin{tabular}{lcccccccccc}
\toprule
\textbf{Config} & \textbf{Threshold} & \textbf{Acc} & \textbf{Prec} & \textbf{Recall} & \textbf{F1} & \textbf{LogLoss} & \textbf{TP} & \textbf{FP} & \textbf{TN} & \textbf{FN} \\
\midrule
C1 & 0.2 & 0.505 & 0.448 & 0.869 & 0.591 & 0.7021 & 630 & 776 & 260 & 95 \\
C2 & 0.26 & 0.445 & 0.424 & 0.97 & 0.59 & 0.6771 & 703 & 956 & 80 & 22 \\
C3 & 0.24 & 0.412 & 0.412 & 1 & 0.584 & 0.6807 & 725 & 1035 & 1 & 0 \\
C4 & 0.22 & 0.413 & 0.412 & 0.997 & 0.583 & 0.6627 & 723 & 1031 & 5 & 2 \\
C5 & 0.22 & 0.507 & 0.451 & 0.919 & 0.605 & 0.6534 & 666 & 810 & 226 & 59 \\
C6 & 0.39 & 0.603 & 0.512 & 0.778 & 0.617 & 0.6302 & 564 & 538 & 498 & 161 \\
C7 & 0.36 & 0.581 & 0.495 & 0.81 & 0.614 & 0.6309 & 587 & 599 & 437 & 138 \\
B1 & 0.27 & 0.537 & 0.466 & 0.865 & 0.606 & 0.6578 & 627 & 718 & 318 & 98 \\
B2 & 0.26 & 0.566 & 0.484 & 0.85 & 0.617 & 0.6626 & 616 & 656 & 380 & 109 \\
B3 & 0.33 & 0.637 & 0.541 & 0.779 & 0.638 & 0.8198 & 565 & 480 & 556 & 160 \\
B4 & 0.28 & 0.409 & 0.41 & 0.994 & 0.581 & 0.6854 & 721 & 1036 & 0 & 4 \\
\bottomrule
\end{tabular}%
}
\end{table*}

\begin{table*}[htbp]
\centering
\caption{Table H+2}%
\label{tab:tabla_h2}
\scriptsize
\resizebox{\textwidth}{!}{%
\begin{tabular}{lcccccccccc}
\toprule
\textbf{Config} & \textbf{Threshold} & \textbf{Acc} & \textbf{Prec} & \textbf{Recall} & \textbf{F1} & \textbf{LogLoss} & \textbf{TP} & \textbf{FP} & \textbf{TN} & \textbf{FN} \\
\midrule
C1 & 0.13 & 0.459 & 0.427 & 0.932 & 0.586 & 0.7122 & 675 & 904 & 132 & 49 \\
C2 & 0.22 & 0.413 & 0.412 & 1 & 0.584 & 0.6816 & 724 & 1033 & 3 & 0 \\
C3 & 0.23 & 0.411 & 0.411 & 0.999 & 0.582 & 0.6914 & 723 & 1036 & 0 & 1 \\
C4 & 0.22 & 0.412 & 0.412 & 1 & 0.583 & 0.6875 & 724 & 1035 & 1 & 0 \\
C5 & 0.24 & 0.512 & 0.452 & 0.877 & 0.597 & 0.6766 & 635 & 770 & 266 & 89 \\
C6 & 0.31 & 0.523 & 0.458 & 0.874 & 0.601 & 0.645 & 633 & 748 & 288 & 91 \\
C7 & 0.31 & 0.528 & 0.462 & 0.881 & 0.606 & 0.6488 & 638 & 744 & 292 & 86 \\
B1 & 0.18 & 0.419 & 0.414 & 0.989 & 0.584 & 0.6674 & 716 & 1014 & 22 & 8 \\
B2 & 0.22 & 0.53 & 0.462 & 0.872 & 0.604 & 0.6842 & 631 & 735 & 301 & 93 \\
B3 & 0.23 & 0.518 & 0.455 & 0.859 & 0.595 & 0.8006 & 622 & 746 & 290 & 102 \\
B4 & 0.28 & 0.411 & 0.411 & 0.997 & 0.582 & 0.6852 & 722 & 1035 & 1 & 2 \\
\bottomrule
\end{tabular}%
}
\end{table*}

\begin{table*}[htbp]
\centering
\caption{Table H+3}%
\label{tab:tabla_h3}
\scriptsize
\resizebox{\textwidth}{!}{%
\begin{tabular}{lcccccccccc}
\toprule
\textbf{Config} & \textbf{Threshold} & \textbf{Acc} & \textbf{Prec} & \textbf{Recall} & \textbf{F1} & \textbf{LogLoss} & \textbf{TP} & \textbf{FP} & \textbf{TN} & \textbf{FN} \\
\midrule
C1 & 0.1 & 0.441 & 0.421 & 0.957 & 0.584 & 0.7172 & 692 & 953 & 83 & 31 \\
C2 & 0.2 & 0.412 & 0.411 & 1 & 0.583 & 0.6809 & 723 & 1034 & 2 & 0 \\
C3 & 0.2 & 0.412 & 0.411 & 1 & 0.583 & 0.698 & 723 & 1035 & 1 & 0 \\
C4 & 0.23 & 0.412 & 0.411 & 0.996 & 0.582 & 0.6932 & 720 & 1032 & 4 & 3 \\
C5 & 0.27 & 0.528 & 0.459 & 0.845 & 0.595 & 0.6808 & 611 & 719 & 317 & 112 \\
C6 & 0.27 & 0.478 & 0.436 & 0.92 & 0.591 & 0.6584 & 665 & 861 & 175 & 58 \\
C7 & 0.26 & 0.486 & 0.44 & 0.929 & 0.598 & 0.6573 & 672 & 854 & 182 & 51 \\
B1 & 0.14 & 0.428 & 0.417 & 0.985 & 0.586 & 0.6642 & 712 & 995 & 41 & 11 \\
B2 & 0.32 & 0.579 & 0.493 & 0.775 & 0.602 & 0.7035 & 560 & 577 & 459 & 163 \\
B3 & 0.23 & 0.501 & 0.444 & 0.842 & 0.581 & 0.7991 & 609 & 764 & 272 & 114 \\
B4 & 0.28 & 0.411 & 0.411 & 0.999 & 0.582 & 0.685 & 722 & 1035 & 1 & 1 \\
\bottomrule
\end{tabular}%
}
\end{table*}

\begin{table*}[htbp]
\centering
\caption{Table H+4}%
\label{tab:tabla_h4}
\scriptsize
\resizebox{\textwidth}{!}{%
\begin{tabular}{lcccccccccc}
\toprule
\textbf{Config} & \textbf{Threshold} & \textbf{Acc} & \textbf{Prec} & \textbf{Recall} & \textbf{F1} & \textbf{LogLoss} & \textbf{TP} & \textbf{FP} & \textbf{TN} & \textbf{FN} \\
\midrule
C1 & 0.1 & 0.441 & 0.421 & 0.954 & 0.584 & 0.7187 & 689 & 949 & 87 & 33 \\
C2 & 0.24 & 0.424 & 0.415 & 0.989 & 0.585 & 0.6767 & 714 & 1005 & 31 & 8 \\
C3 & 0.19 & 0.411 & 0.411 & 1 & 0.582 & 0.7009 & 722 & 1035 & 1 & 0 \\
C4 & 0.28 & 0.429 & 0.418 & 0.986 & 0.587 & 0.6776 & 712 & 993 & 43 & 10 \\
C5 & 0.26 & 0.519 & 0.455 & 0.867 & 0.597 & 0.6696 & 626 & 750 & 286 & 96 \\
C6 & 0.33 & 0.535 & 0.463 & 0.835 & 0.596 & 0.6562 & 603 & 698 & 338 & 119 \\
C7 & 0.3 & 0.513 & 0.451 & 0.868 & 0.594 & 0.6575 & 627 & 762 & 274 & 95 \\
B1 & 0.33 & 0.582 & 0.494 & 0.776 & 0.604 & 0.6517 & 560 & 573 & 463 & 162 \\
B2 & 0.16 & 0.494 & 0.443 & 0.896 & 0.593 & 0.7046 & 647 & 814 & 222 & 75 \\
B3 & 0.12 & 0.443 & 0.42 & 0.932 & 0.579 & 0.8149 & 673 & 930 & 106 & 49 \\
B4 & 0.28 & 0.412 & 0.411 & 1 & 0.583 & 0.6847 & 722 & 1034 & 2 & 0 \\
\bottomrule
\end{tabular}%
}
\end{table*}

\begin{table*}[htbp]
\centering
\caption{Table H+5}%
\label{tab:tabla_h5}
\scriptsize
\resizebox{\textwidth}{!}{%
\begin{tabular}{lcccccccccc}
\toprule
\textbf{Config} & \textbf{Threshold} & \textbf{Acc} & \textbf{Prec} & \textbf{Recall} & \textbf{F1} & \textbf{LogLoss} & \textbf{TP} & \textbf{FP} & \textbf{TN} & \textbf{FN} \\
\midrule
C1 & 0.14 & 0.462 & 0.428 & 0.917 & 0.583 & 0.7212 & 662 & 886 & 149 & 60 \\
C2 & 0.27 & 0.479 & 0.437 & 0.924 & 0.593 & 0.664 & 667 & 861 & 174 & 55 \\
C3 & 0.24 & 0.41 & 0.41 & 0.997 & 0.581 & 0.6851 & 720 & 1035 & 0 & 2 \\
C4 & 0.28 & 0.441 & 0.421 & 0.961 & 0.585 & 0.6623 & 694 & 955 & 80 & 28 \\
C5 & 0.17 & 0.467 & 0.433 & 0.957 & 0.596 & 0.6706 & 691 & 905 & 130 & 31 \\
C6 & 0.31 & 0.546 & 0.471 & 0.848 & 0.605 & 0.6506 & 612 & 688 & 347 & 110 \\
C7 & 0.26 & 0.508 & 0.451 & 0.9 & 0.601 & 0.6538 & 650 & 792 & 243 & 72 \\
B1 & 0.28 & 0.557 & 0.477 & 0.83 & 0.606 & 0.6509 & 599 & 656 & 379 & 123 \\
B2 & 0.21 & 0.53 & 0.461 & 0.845 & 0.597 & 0.7114 & 610 & 713 & 322 & 112 \\
B3 & 0.1 & 0.435 & 0.417 & 0.939 & 0.577 & 0.83 & 678 & 949 & 86 & 44 \\
B4 & 0.28 & 0.411 & 0.411 & 1 & 0.583 & 0.685 & 722 & 1034 & 1 & 0 \\
\bottomrule
\end{tabular}%
}
\end{table*}

\clearpage
\noindent\begin{minipage}{\textwidth}
\centering
\captionof{table}{Table H+6}%
\label{tab:tabla_h6}
\scriptsize
\resizebox{\textwidth}{!}{%
\begin{tabular}{lcccccccccc}
\toprule
\textbf{Config} & \textbf{Threshold} & \textbf{Acc} & \textbf{Prec} & \textbf{Recall} & \textbf{F1} & \textbf{LogLoss} & \textbf{TP} & \textbf{FP} & \textbf{TN} & \textbf{FN} \\
\midrule
C1 & 0.12 & 0.451 & 0.424 & 0.94 & 0.585 & 0.7231 & 679 & 921 & 113 & 43 \\
C2 & 0.3 & 0.536 & 0.465 & 0.843 & 0.599 & 0.6604 & 609 & 702 & 332 & 113 \\
C3 & 0.18 & 0.412 & 0.411 & 1 & 0.583 & 0.6908 & 722 & 1033 & 1 & 0 \\
C4 & 0.26 & 0.42 & 0.414 & 0.986 & 0.583 & 0.6735 & 712 & 1009 & 25 & 10 \\
C5 & 0.21 & 0.504 & 0.449 & 0.911 & 0.602 & 0.674 & 658 & 807 & 227 & 64 \\
C6 & 0.27 & 0.492 & 0.444 & 0.928 & 0.6 & 0.6574 & 670 & 840 & 194 & 52 \\
C7 & 0.29 & 0.521 & 0.457 & 0.877 & 0.601 & 0.6626 & 633 & 753 & 281 & 89 \\
B1 & 0.26 & 0.538 & 0.466 & 0.866 & 0.606 & 0.6541 & 625 & 715 & 319 & 97 \\
B2 & 0.21 & 0.528 & 0.459 & 0.831 & 0.591 & 0.7246 & 600 & 707 & 327 & 122 \\
B3 & 0.1 & 0.438 & 0.419 & 0.939 & 0.579 & 0.8536 & 678 & 942 & 92 & 44 \\
B4 & 0.28 & 0.412 & 0.411 & 0.999 & 0.583 & 0.6856 & 721 & 1032 & 2 & 1 \\
\bottomrule
\end{tabular}%
}

\vspace{0.8em}

\captionof{table}{Table H+7}%
\label{tab:tabla_h7}
\scriptsize
\resizebox{\textwidth}{!}{%
\begin{tabular}{lcccccccccc}
\toprule
\textbf{Config} & \textbf{Threshold} & \textbf{Acc} & \textbf{Prec} & \textbf{Recall} & \textbf{F1} & \textbf{LogLoss} & \textbf{TP} & \textbf{FP} & \textbf{TN} & \textbf{FN} \\
\midrule
C1 & 0.12 & 0.449 & 0.423 & 0.936 & 0.583 & 0.7313 & 676 & 921 & 112 & 46 \\
C2 & 0.25 & 0.435 & 0.42 & 0.983 & 0.589 & 0.6754 & 710 & 979 & 54 & 12 \\
C3 & 0.23 & 0.42 & 0.414 & 0.988 & 0.583 & 0.7049 & 713 & 1009 & 24 & 9 \\
C4 & 0.22 & 0.413 & 0.411 & 0.994 & 0.582 & 0.6971 & 718 & 1027 & 6 & 4 \\
C5 & 0.16 & 0.468 & 0.433 & 0.95 & 0.595 & 0.6901 & 686 & 897 & 136 & 36 \\
C6 & 0.25 & 0.455 & 0.427 & 0.949 & 0.589 & 0.6715 & 685 & 919 & 114 & 37 \\
C7 & 0.23 & 0.446 & 0.422 & 0.945 & 0.584 & 0.6728 & 682 & 933 & 100 & 40 \\
B1 & 0.2 & 0.436 & 0.421 & 0.992 & 0.591 & 0.6682 & 716 & 983 & 50 & 6 \\
B2 & 0.12 & 0.46 & 0.428 & 0.928 & 0.586 & 0.7409 & 670 & 895 & 138 & 52 \\
B3 & 0.1 & 0.437 & 0.418 & 0.938 & 0.578 & 0.8557 & 677 & 943 & 90 & 45 \\
B4 & 0.28 & 0.411 & 0.411 & 0.999 & 0.583 & 0.686 & 721 & 1032 & 1 & 1 \\
\bottomrule
\end{tabular}%
}
\end{minipage}

\clearpage 
\selectlanguage{spanish} 

\twocolumn[{%
  \centering
  {\LARGE \bf De la estacionalidad a la semántica: benchmarking de un sistema híbrido de predicción probabilística para bloqueos de vías en Bolivia} \\[1.5em]
  
  \begin{minipage}[t]{0.46\textwidth}
    \centering
    \textbf{Rodrigo Vargas Sainz} \\
    \small Universidad Privada de Santa Cruz de la Sierra \\
    \scriptsize \texttt{rodrigovargas@upsa.edu.bo} / \texttt{rodrivs@mit.edu}
  \end{minipage}
  \hfill
  \begin{minipage}[t]{0.46\textwidth}
    \centering
    \textbf{Christian Berón Curti} \\
    \small Universidad Privada de Santa Cruz de la Sierra \\
    \scriptsize \texttt{cberon@mit.edu} / \texttt{cberon@gmail.com}
  \end{minipage} \\[2em]
  
  \date{}
  
  \begin{abstract}
    \justifying%
    Los bloqueos de carreteras en Bolivia son un fenómeno de conflicto social con impactos económicos devastadores, estimados en pérdidas equivalentes al 4\% del Producto Interno Bruto nacional. A pesar de su recurrencia e impacto, existe una carencia de sistemas predictivos locales para anticipar estos eventos en la toma de decisiones logísticas. Este artículo presenta un sistema híbrido de pronóstico probabilístico que integra la descomposición de series temporales (Prophet) con técnicas de procesamiento del lenguaje natural (NLP) aplicadas a un corpus de seis años de cobertura periodística boliviana. La metodología emplea embeddings semánticos vectoriales y modelos de clasificación zero-shot para capturar señales de escalada discursiva antes de la materialización de los bloqueos. Utilizando un esquema de validación walk-forward de ventana expansiva aplicado a lo largo de 1,762 días y siete horizontes de pronóstico (H+1 a H+7), se compararon siete configuraciones internas y cuatro benchmarks externos, incluyendo SARIMA y LightGBM\@. Los resultados demuestran que la configuración híbrida (Prophet + NLP, C6) supera de manera consistente a los modelos puramente estadísticos, alcanzando un AUC-ROC de 0.677 en H+1 y reduciendo el Brier Score en un 10.9\% en relación con el modelo temporal de referencia (0.220 frente a 0.247), manteniendo una reducción de error estadísticamente significativa en todos los horizontes evaluados ($p < 0.02$). Esta investigación valida que la integración de señales semánticas de noticias permite la detección de picos de tensión social no capturados por la inercia histórica, proporcionando una herramienta técnica para la gestión de riesgos en corredores de transporte críticos.
    
    \vspace{0.4cm}
    \noindent\textbf{Palabras clave:} Pronóstico de series temporales, Procesamiento de Lenguaje Natural, Bloqueos de carreteras, Bolivia, Prophet, XGBoost, Modelos híbridos.
    \vspace{0.6cm}
  \end{abstract}
}]

\section{Introducción}

Los bloqueos de carreteras en Bolivia tienen una trayectoria histórica que excede su función original como un mecanismo disruptivo de negociación sectorial. Aunque los bloqueos tienen incluso orígenes coloniales, existen registros más contemporáneos. Podríamos remontarnos a la ``Guerra del Agua'' (2000) o la ``Guerra del Gas'' (2003)~\cite{1}, eventos históricos que establecieron un modelo de movilización capaz de reorientar la política económica en el país~\cite{2}. Durante el período de gobierno del Movimiento al Socialismo (MAS), los bloqueos experimentaron un proceso de institucionalización que incorporó a sectores sociales no tradicionales en el debate político ante organismos oficiales~\cite{3}. Sin embargo, la literatura reciente respecto a estos eventos documenta una transformación de este instrumento: de formas de diálogo social, han degenerado en un recurso de coacción interna dentro de las organizaciones sociales o poblaciones, sirviendo como una herramienta de presión política aparentemente orientada hacia objetivos privados que se entrelazan con demandas colectivas~\cite{4}.

Una parte de los estudios revisados del período 2020--2025 atribuye la intensificación de los conflictos viales a la división interna del MAS entre las facciones identificadas como ``arcista'' y ``evista'' (grupos de apoyo al expresidente Luis Arce Catacora versus grupos de apoyo al expresidente Evo Morales Ayma), en las que las carreteras se convirtieron en arenas de disputa por diversas razones, que van desde el control de la sigla partidaria (MAS) hasta la pugna sobre quién decide las políticas del país: el partido político o el gobierno~\cite{5}. Este patrón de conflicto se intensificó con la posesión presidencial de Rodrigo Paz Pereira en 2025, cuya eliminación de los subsidios a los hidrocarburos —que elevó el precio de la gasolina en un 163\%— provocó una ola de protestas y bloqueos a escala nacional~\cite{6}.

El impacto económico y social de estos episodios ha sido documentado de manera consistente en informes técnicos de diversas organizaciones y artículos periodísticos. Se estima que un mes de bloqueos continuos genera pérdidas equivalentes al 4\% del Producto Interno Bruto nacional~\cite{7}. A nivel sectorial, la industria nacional reporta pérdidas millonarias, mientras que el sector turístico enfrenta una crisis que pone en riesgo miles de empleos~\cite{8}. Los efectos sociales incluyen hasta la triplicación de los precios de los alimentos básicos debido al desabastecimiento y casos documentados de mortalidad asociados a la imposibilidad de trasladar pacientes o suministrar oxígeno médico~\cite{1}. Los estudios de geografía política también han identificado un patrón de concentración territorial de los puntos de bloqueo en municipios con una fuerte presencia de organizaciones específicas, como en la región del Chapare~\cite{9}.

El fenómeno de los bloqueos de carreteras ha impulsado un debate jurisprudencial no resuelto respecto a su estatus legal: si bien la Constitución Política del Estado boliviano reconoce el derecho a la libertad de reunión y de expresión, no respalda el derecho a bloquear las vías públicas. Han existido y continúan existiendo diversas propuestas legislativas, como el proyecto de ley que propone sanciones por la interrupción del libre tránsito, las cuales han generado controversia entre analistas que advierten sobre los riesgos de criminalizar la protesta social y aquellos que sostienen la necesidad de garantizar el derecho a la libre circulación de los ciudadanos no involucrados en los conflictos~\cite{10}.

A pesar de la magnitud de este impacto económico y social, y de la consolidación de los bloqueos como un evento recurrente, no existen sistemas predictivos que permitan anticipar la ocurrencia de estos eventos con la suficiente antelación para la toma de decisiones logísticas y de gestión de riesgos más allá de la intuición y la lectura de noticias por parte de empresas u organizaciones interesadas. Aunque la literatura sobre detección temprana de conflictos ha validado la utilidad de las fuentes textuales en contextos de alta visibilidad, persiste una brecha analítica con respecto a su aplicabilidad en escenarios de conflicto recurrente de baja intensidad. El caso de los bloqueos de carreteras en Bolivia, caracterizado por su especificidad territorial, presenta un desafío que los enfoques globales como GDELT (Global Database of Events, Language, and Tone)~\cite{11} no pueden resolver por sí solos.

El presente trabajo aborda esta problemática mediante el desarrollo de un sistema híbrido de pronóstico probabilístico que combina la descomposición de series temporales (capturando la estructura cíclica de los bloqueos causados por conflictos sociales) con técnicas de procesamiento del lenguaje natural aplicadas a un corpus de seis años de cobertura periodística (capturando señales de escalada discursiva previas a la materialización de los eventos). A través de un estudio de ablación de siete configuraciones internas (C1--C7) y cuatro benchmarks externos independientes (regresión logística regularizada, LightGBM, SARIMA univariado y XGBoost con características autorregresivas) evaluados bajo un esquema de validación walk-forward durante 1,762 días fuera de muestra, cuantificamos si la contribución semántica de la prensa es estadísticamente significativa y si esta ganancia es atribuible a la arquitectura de información propuesta o al algoritmo de aprendizaje empleado.

El resto de este artículo se organiza de la siguiente manera: la Sección 2 revisa la literatura relacionada sobre la detección de conflictos sociales mediante NLP y el pronóstico híbrido de series temporales. La Sección 3 describe los datos y la metodología propuesta. La Sección 4 presenta los resultados del estudio de ablación y las pruebas de significancia estadística. La Sección 5 discute los hallazgos y sus limitaciones. Finalmente, la Sección 6 concluye con las contribuciones de este trabajo y las líneas de investigación futuras.

\section{Trabajo Relacionado}

La anticipación de la inestabilidad sociopolítica ha pasado progresivamente de enfoques basados en el juicio de expertos hacia sistemas de alerta temprana respaldados por grandes volúmenes de datos no estructurados. La disponibilidad de bases de datos de cobertura global como GDELT permite el monitoreo continuo de eventos de inestabilidad a partir de fuentes periodísticas~\cite{12}. El procesamiento del lenguaje natural aplicado a este dominio ha evolucionado desde la detección de palabras clave y la aplicación de la Teoría de Grafos hacia la extracción estructurada de eventos mediante ontologías especializadas, destacando el marco CAMEO (Conflict and Mediation Event Observations)~\cite{13}, adoptado por sistemas operacionales como EMBERS, los cuales son capaces de generar alertas estructuradas que especifican el actor, la ubicación, la temporalidad y el tipo de evento a partir de fuentes abiertas como la red social X, blogs y prensa digital~\cite{14}. Estos sistemas han reportado una capacidad de anticipación (tiempo de antelación) superior a la cobertura de noticias tradicional en aplicaciones de América Latina, aunque persisten limitaciones documentadas~\cite{14}.

Trabajos previos sobre la predicción del malestar civil a partir de redes sociales han explorado enfoques complementarios a la extracción estructurada de eventos. Un grupo de investigadores propuso un modelo basado en la detección de cascadas de actividad en Twitter para predecir la ocurrencia de protestas en varios países de América Latina, bajo el supuesto de que el surgimiento de cascadas de actividad a gran escala constituye un precursor observable de la movilización física~\cite{15}. Weber et al.\ evaluaron predictores derivados de la teoría de participación en protestas utilizando datos de Twitter durante la Primavera Árabe egipcia, encontrando respaldo significativo únicamente para el incremento en el volumen de expresiones que anticipan explícitamente eventos de protesta futuros, sin evidencia significativa para las hipótesis relacionadas con el volumen general de actividad política en redes sociales~\cite{16}. En una línea diferente, Hlatshwayo y Redl, utilizando un índice de cientos de indicadores socioeconómicos, entrenaron modelos basados en árboles de decisión para predecir disturbios con un año de anticipación. Descubrieron que, además del historial de disturbios previos, la inflación de los precios de los alimentos y la penetración de la telefonía móvil eran los predictores más robustos~\cite{17}.

Investigaciones recientes han incorporado modelos de lenguaje de gran tamaño (LLM) como un mecanismo de razonamiento contextual para la dinámica de conflictos~\cite{18}. De manera similar, el modelo ExoLLM transfiere representaciones aprendidas del dominio del lenguaje natural al pronóstico numérico, integrando descripciones textuales de los eventos~\cite{19}.

La literatura sobre pronóstico de series temporales aplicada a fenómenos de conflicto ha identificado ventajas consistentes en arquitecturas híbridas que combinan la robustez estadística con la capacidad de aprendizaje no lineal. La combinación ARIMA-LSTM constituye la arquitectura dominante en este aspecto: el componente ARIMA captura la tendencia lineal y la estructura estacional de la serie, mientras que una red de Memoria a Corto y Largo Plazo (LSTM) se entrena con los residuos del ajuste lineal para capturar dinámicas no lineales y picos abruptos de actividad no explicados por el componente estadístico~\cite{20}. Trabajos posteriores han expandido este principio mediante mecanismos de combinación dinámica que permiten al sistema adaptar el peso relativo de cada componente según el régimen de volatilidad observado, optimizando la robustez del pronóstico final~\cite{21}.

Más allá de las variables puramente endógenas, los modelos recientes incorporan variables exógenas adicionales, tales como indicadores económicos (inflación de alimentos, tipos de cambio) y señales de comportamiento digital, incluyendo el uso de redes de anonimización como Tor como proxy para la evasión de la censura ante una represión inminente, así como cascadas de actividad en redes sociales como las descritas previamente.

Al analizar la literatura actual sobre el pronóstico de conflictos sociales, emergen dos tendencias claras: Primero, una transición metodológica desde modelos estadísticos univariados hacia arquitecturas híbridas de descomposición temporal. Aunque el conflicto sigue estructuras estacionales predecibles, los picos abruptos de actividad dependen fundamentalmente de variables exógenas o del suavizado de picos estacionales. Por lo tanto, los sistemas actuales se enfocan en aislar componentes deterministas mediante modelos estadísticos robustos, reservando el procesamiento del lenguaje natural (NLP) para modelar residuales e intensidades latentes. Segundo, una asimetría geográfica: los estudios empíricos se concentran en contextos en idioma inglés o en eventos de visibilidad internacional, dejando una brecha en entornos de conflicto recurrente y baja visibilidad mediática, como es el caso de Bolivia.

No existe en la literatura actual ningún sistema aplicado específicamente a los bloqueos de carreteras en Bolivia, ni una evaluación que cuantifique el valor marginal proporcionado por las representaciones semánticas de la prensa sobre modelos de descomposición temporal como Prophet~\cite{22}, ni una que compare dicha arquitectura con benchmarks independientes de aprendizaje automático y series temporales. La presente propuesta aborda ambas brechas: combina intencionadamente la interpretabilidad de la descomposición temporal con la precisión local de modelos encoder-only y gradient boosting calibrado, y valida la contribución del componente semántico frente a cuatro líneas de base externas---regresión logística regularizada (B1), LightGBM (B2), SARIMA univariado (B3) y XGBoost con características autorregresivas (B4)---todas evaluadas bajo el mismo esquema walk-forward, con hiperparámetros optimizados sobre una ventana de ajuste independiente del período de evaluación final para garantizar que la comparación esté libre de sesgos de selección de modelos.

\section{Datos y Metodología}

\subsection{Fuentes de datos}

Para este estudio, se integraron dos fuentes de información independientes con diferentes horizontes temporales. La primera fuente es un archivo histórico de un portal de noticias boliviano que funciona como un agregador de noticias que replica y centraliza el contenido producido por diversos medios digitales nacionales. Este archivo contiene 916,110 titulares de noticias sin procesar (titulares, enlaces, marcas de tiempo y etiquetas), que abarcan desde 2008 hasta 2026. Para el período de análisis (2020--2026), el corpus filtrado temáticamente comprende 386,884 artículos distribuidos a lo largo de 2,137 días, con un promedio de 181 noticias relevantes por día. Los 517,426 artículos restantes corresponden a la cobertura previa a 2020, un período sin registros de eventos viales en la ABC, y fueron descartados de la ventana de evaluación. La segunda fuente proviene de los registros oficiales de la Administradora Boliviana de Carreteras (ABC), los cuales documentan 37,903 eventos del estado de carreteras entre el 13 de agosto de 2020 y el 13 de junio de 2026. Para nuestra investigación, restringimos el análisis al período compartido (2020--2026), filtrando los datos de la ABC para incluir exclusivamente interrupciones debidas a conflictos sociales en seis corredores estratégicos para el transporte nacional (Rutas 1, 4, 5, 6, 7 y 25). La agregación de la variable objetivo a nivel nacional responde a una restricción intrínseca de la fuente de datos textuales: más del 90\% de los titulares de prensa omiten la nomenclatura específica de la ruta vial (p.\ ej., Ruta 4 o Ruta 1), limitándose a reportar la intencionalidad del conflicto social. Forzar una desagregación espacial a nivel de ruta introduciría un sesgo de datos faltantes y reduciría drásticamente el poder estadístico del conjunto de datos. Por lo tanto, el marco híbrido se formula como un proxy del riesgo sistémico nacional.

\begin{figure*}[htbp]
\centering
\includegraphics[width=\textwidth]{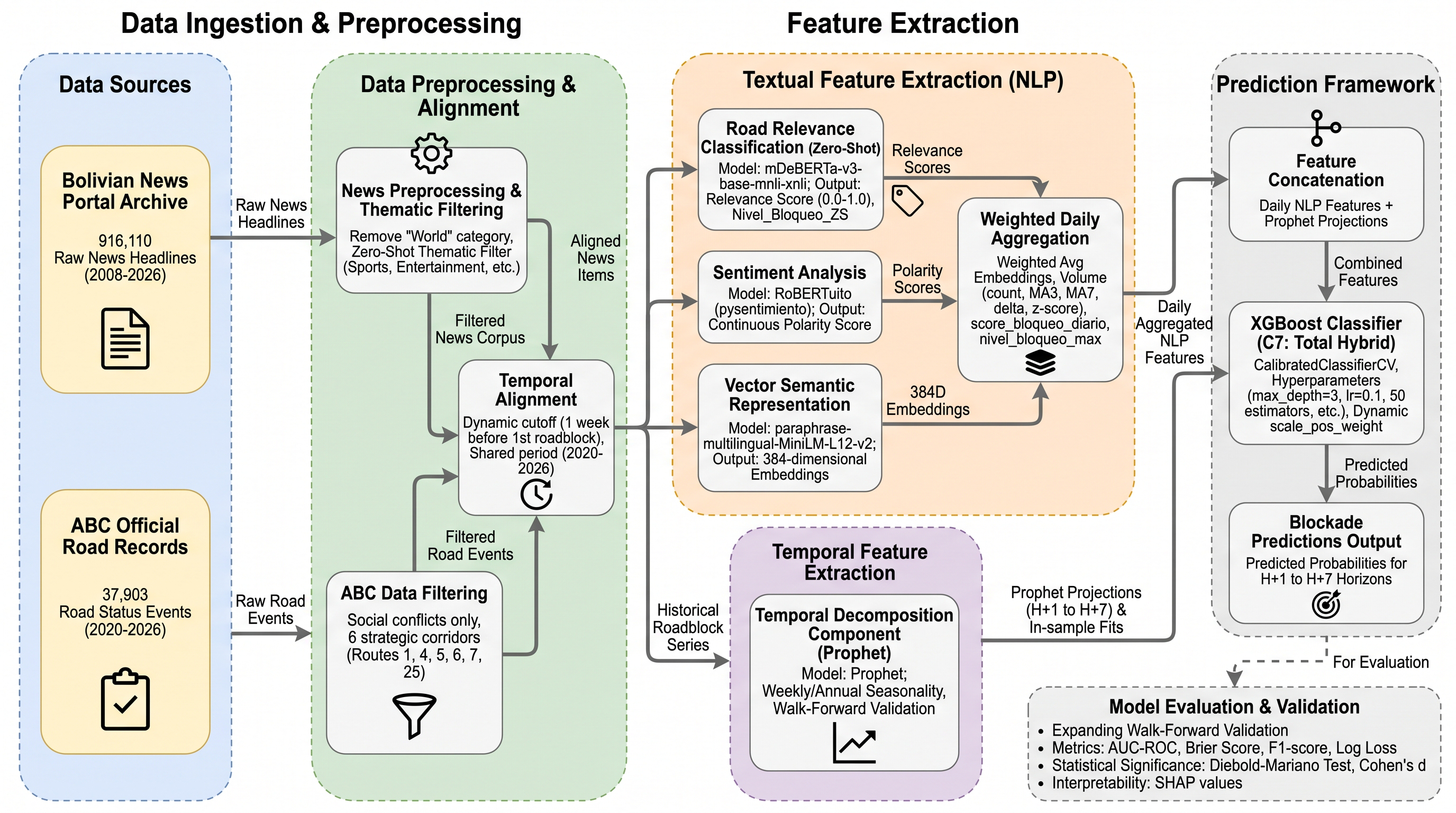}
\caption{Resultado final del proceso de integración y modelado.}%
\label{es:fig:final_output}
\end{figure*}

\subsection{Preprocesamiento y filtrado temático}

El corpus de noticias se depuró en dos fases. En primer lugar, se eliminó la categoría ``Mundo'' debido a su falta de relevancia para el conflicto local. En segundo lugar, aplicamos un filtro basado en clasificación temática zero-shot (detallada en la Sección 3.3) para excluir contenido no relacionado con el fenómeno, como deportes, entretenimiento, horóscopo y moda. Tras este proceso, el corpus resultante se ajustó a 903,602 artículos.

Para alinear temporalmente ambas fuentes, se adoptó un criterio de corte dinámico: el inicio de la serie temporal se fijó una semana antes de la fecha del primer bloqueo registrado en la base de datos de la ABC\@. Este margen garantiza que el modelo cuente con el contexto mediático previo necesario sin incluir períodos de inactividad que no aportan una señal predictiva para el fenómeno estudiado.

\subsection{Extracción de características textuales}

Se aplicaron tres procesos de extracción de características a cada noticia filtrada utilizando modelos de lenguaje preentrenados:

\noindent\textbf{Clasificación de relevancia vial (zero-shot).} Se utilizó un clasificador multilingüe zero-shot (\texttt{mDeBERTa-v3-base-mnli-xnli}) para asignar a cada titular una puntuación de relevancia continua con respecto a cuatro niveles de intensidad de conflicto vial:

\noindent{}-- Bloqueo físico confirmado (peso de referencia 1.0)\par
\noindent{}-- Anuncio de marcha o bloqueo (0.60)\par
\noindent{}-- Paro o demanda sindical (0.20)\par
\noindent{}-- Contenido no relacionado (0.0).\par
Adicionalmente, se registró para cada noticia el nivel discreto con la mayor probabilidad (\texttt{Nivel\_Bloqueo\_ZS}).

\noindent\textbf{Análisis de sentimiento.} Se aplicó un modelo de análisis de sentimiento en español (\texttt{RoBERTuito}, a través de la librería \texttt{pysentimiento}) para obtener una puntuación de polaridad continua por noticia.

\noindent\textbf{Representación semántica vectorial.} Cada titular se transformó en un vector de 384 dimensiones utilizando el modelo de embeddings de oraciones multilingüe \texttt{paraphrase-multilingual-MiniLM-L12-v2}. Se realizó una evaluación preliminar sobre la reducción de esta dimensionalidad a un espacio latente de baja dimensión ($d=10$) mediante el Análisis de Componentes Principales (PCA). Sin embargo, las pruebas experimentales demostraron una degradación sistemática en la capacidad discriminativa del clasificador (AUC-ROC), atribuida a la pérdida masiva de varianza explicada inherente a una compresión tan agresiva de embeddings semánticos densos. En consecuencia, se decidió preservar las 384 dimensiones originales para maximizar la capacidad expresiva y salvaguardar los límites de decisión del modelo.

\subsection{Agregación diaria ponderada}

Las noticias individuales se agregaron a nivel diario utilizando un esquema de ponderación basado en la puntuación de relevancia vial descrita en 3.3. Para cada día $d$, el vector semántico agregado se calculó como el promedio ponderado de los embeddings de todas las noticias publicadas en ese día, utilizando la puntuación de relevancia vial como peso cuando esta fuera positiva, y un peso residual diferenciado (0.05 para noticias clasificadas explícitamente como irrelevantes, 0.30 para noticias sin clasificación zero-shot disponible) en caso contrario. Esta ponderación amplifica la contribución de las noticias con contenido semánticamente relacionado con un conflicto vial activo en comparación con la cobertura mediática de fondo no relacionada.

Adicionalmente, el volumen absoluto de noticias, sus medias móviles de 3 y 7 días, su variación diaria (delta) y su puntuación z normalizada se calcularon como variables diarias, junto con el promedio diario de la puntuación de relevancia vial (\texttt{score\_bloqueo\_diario}) y el nivel máximo de escalada observado en el día (\texttt{nivel\_bloqueo\_max}).

\subsection{Componente de descomposición temporal}

El modelo Prophet~\cite{22} se empleó para capturar la estructura estacional del fenómeno, incorporando componentes de estacionalidad semanal y anual. Para evitar la fuga de información temporal, se ajustó un único modelo Prophet en cada paso del esquema de validación walk-forward (Sección 3.7) sobre el historial disponible hasta el día de evaluación, generando simultáneamente los valores ajustados dentro de la muestra (utilizados como variable predictora en el conjunto de entrenamiento) y las proyecciones fuera de la muestra para los siete horizontes de pronóstico evaluados.

\subsection{Arquitectura de ablación de siete niveles}

Con el fin de aislar la contribución incremental de cada fuente de información, se diseñaron siete configuraciones experimentales, cada una entrenada utilizando un clasificador XGBoost calibrado (escalado de Platt a través de \texttt{CalibratedClassifierCV}) con los siguientes hiperparámetros uniformes: profundidad máxima de 3, tasa de aprendizaje de 0.1, 50 estimadores, submuestreo de columnas por árbol de 0.3 (regularización implícita dada la alta dimensionalidad de los embeddings), submuestreo de filas de 0.8 y un peso dinámico de la clase positiva (\texttt{scale\_pos\_weight}) recalculado en cada ventana de entrenamiento como la razón entre observaciones negativas y positivas, para compensar el desequilibrio moderado de clases (prevalencia histórica $\approx$ 38--41\%).

Las siete configuraciones fueron:

-C1 (Prophet Puro): solo la proyección de descomposición temporal, recalibrada mediante regresión logística (escalado de Platt) sobre los valores ajustados dentro de la muestra.

-C2 (Volumen): solo las variables de volumen de prensa (conteo diario, medias móviles, delta, puntuación z).

-C3 (NLP Puro): solo el vector semántico agregado de 384 dimensiones, sin los componentes temporal o de volumen.

-C4 (Volumen + Zero-Shot + NLP): integración de volumen, puntuación de relevancia vial y semántica, sin el componente temporal.

-C5 (Prophet + Zero-Shot): componente temporal combinado con volumen y puntuación de relevancia vial.

-C6 (Prophet + NLP): componente temporal combinado con volumen y semántica vectorial completa.

-C7 (Híbrido Total): integración de las cuatro fuentes de información (temporal, volumen, puntuación de relevancia vial, semántica).

Para determinar si la ganancia observada en las configuraciones C5--C7 es atribuible a la arquitectura de información propuesta o al algoritmo de aprendizaje empleado, se evaluaron adicionalmente cuatro benchmarks externos bajo el mismo esquema walk-forward y los mismos horizontes:

\noindent{}-- B1 (Regresión Logística Regularizada) C=0.01.\par
\noindent{}-- B2 (LightGBM) 100 estimadores, profundidad 3, tasa de aprendizaje 0.05\par
\noindent{}-- B3 (SARIMA Univariado) de orden (1,1,1)$\times$ (1,1,1,7), reajustado en cada paso del walk-forward sobre la serie \texttt{hubo\_bloqueo}\par
\noindent{}-- B4 (XGBoost) con las mismas características que C7 más rezagos de la variable objetivo de 1, 2, 3 y 7 días.\par
Los hiperparámetros de B1, B2 y B4 se seleccionaron mediante búsqueda en rejilla sobre una ventana de ajuste independiente (del 15 de agosto de 2021 al 31 de diciembre de 2023), excluyendo el período de evaluación final (del 1 de enero de 2024 al 11 de junio de 2026) para evitar el sesgo de selección de modelos. Los benchmarks reciben las mismas características que C7, garantizando que cualquier diferencia observada refleje la capacidad del algoritmo y no la disponibilidad de información.

\subsection{Esquema de validación: walk-forward expandente}

La evaluación se llevó a cabo utilizando un esquema de validación walk-forward de ventana expansiva, considerado el estándar metodológico para prevenir la fuga de información en problemas de pronóstico de series temporales~\cite{23}. Para cada día $t$ del período de evaluación (del 15 de agosto de 2021 al 11 de junio de 2026; 1,762 días), las siete configuraciones se entrenaron utilizando exclusivamente información disponible hasta el día $t-1$, y se generaron predicciones para los siete horizontes de $t+1$ a $t+7$. La variable objetivo de pronóstico ($Y$) en cada horizonte $h$ se construyó desplazando la variable binaria de ocurrencia de bloqueos, garantizando que la variable predictora \texttt{inercia\_prophet} utilizada en el entrenamiento correspondiera, para cada observación histórica, a la proyección de Prophet generada para el día de impacto correspondiente a ese horizonte, replicando exactamente la estructura de información disponible en el momento de la inferencia real.

\subsection{Métricas de evaluación}

Se reportan cuatro familias de métricas. Discriminación: área bajo la curva ROC (AUC-ROC). Calibración: Brier Score y Brier Skill Score (BSS) con respecto a la climatología histórica del problema, definida como $p(1-p)$ donde $p$ es la prevalencia global de la clase positiva. Clasificación binaria: precisión, exhaustividad (recall), F1-score y log loss, calculados sobre el umbral de decisión que maximiza el F1-score dentro del conjunto fuera de muestra para cada horizonte. Significancia estadística: prueba de Diebold–Mariano (Diebold \& Mariano, 1995) con corrección de varianza de Newey–West para horizontes múltiples, aplicada a siete comparaciones pareadas de interés científico, junto con el tamaño del efecto ($d$ de Cohen) sobre el diferencial del error cuadrático entre configuraciones. Adicionalmente, este trabajo incluye curvas de calibración (diagramas de confiabilidad) para las configuraciones establecidas y los benchmarks en horizontes seleccionados (H+1, H+4 y H+7), brindando la oportunidad de evaluar la coherencia entre las probabilidades emitidas y las frecuencias correspondientes. La interpretabilidad del modelo C7 también se examina utilizando valores SHAP (SHapley Additive exPlanations), calculados sobre el clasificador XGBoost base antes de la calibración de Platt, con las 384 dimensiones de los embeddings agregadas como un único grupo para facilitar la lectura.

\section{Resultados y Discusión}

\subsection{Evaluación de rendimiento predictivo}

La evaluación de las siete configuraciones experimentales bajo el esquema de validación walk-forward evidencia una división entre dos grupos de modelos. Las configuraciones que incorporan el componente de descomposición temporal (C1, C5, C6, C7) superan a aquellas que no lo hacen (C2, C3, C4) en AUC-ROC y Brier Score en casi todos los horizontes evaluados. Dentro del primer grupo, la incorporación de señales semánticas de noticias sobre la línea de base estacional (C6: Prophet+NLP) produce una mejora consistente con respecto a Prophet puro (C1) a lo largo de los siete horizontes, tanto en capacidad discriminativa como en calibración probabilística.

\begin{figure*}[htbp]
\centering
\includegraphics[width=\textwidth]{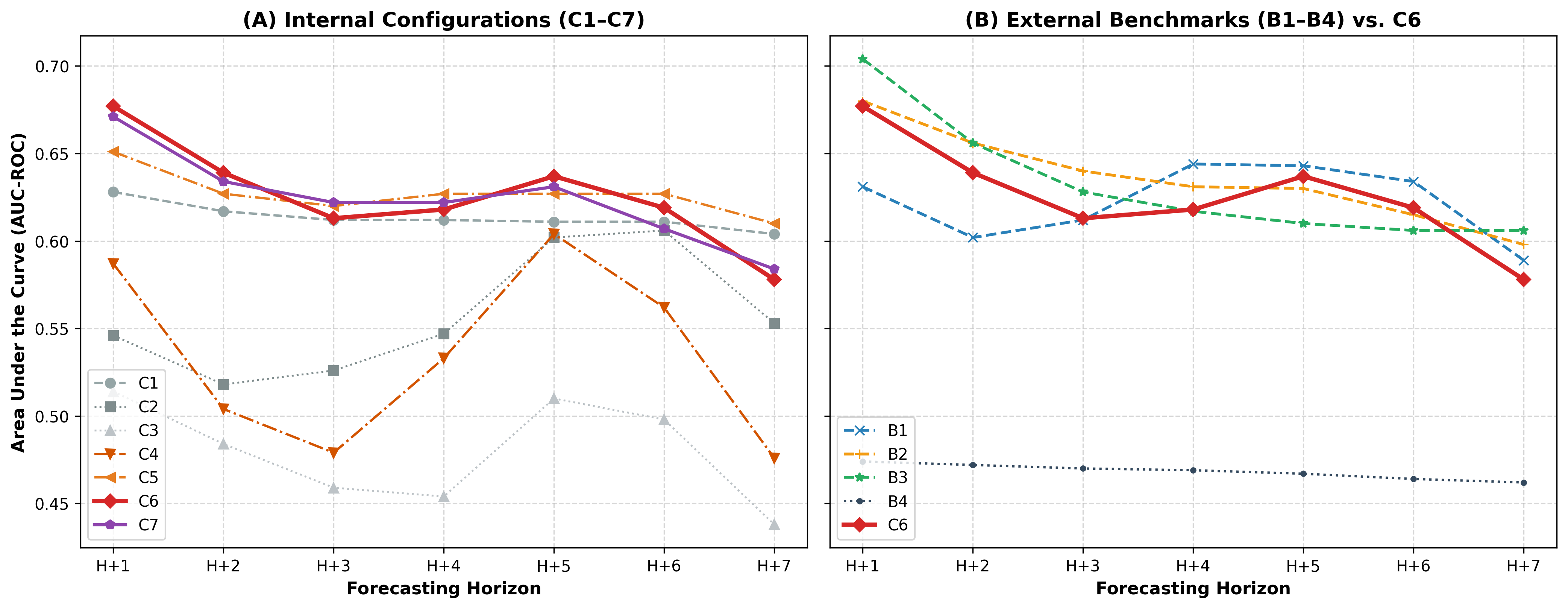}
\caption{Area bajo la curva (AUC) por horizonte de pronóstico}%
\label{es:fig:auc_por_horizonte}
\end{figure*}

La comparación entre C6 y C7 (Híbrido Total) revela que no existe una diferencia estadísticamente significativa en la mayoría de los horizontes. Esta equivalencia ocurre porque la puntuación de relevancia zero-shot ya está incorporada implícitamente en C6 como el mecanismo de ponderación durante la agregación diaria de embeddings (Sección 3.4). En consecuencia, añadir la puntuación zero-shot como una característica tabular explícita en C7 proporciona información redundante, confirmando a C6 como la arquitectura más parsimoniosa.

La evaluación de los cuatro benchmarks externos sobre el mismo período de walk-forward revela un patrón distinto que depende tanto del horizonte como de la métrica considerada. En términos de discriminación (AUC-ROC), el benchmark más competitivo en H+1 resulta ser B3 (SARIMA univariado), con AUC=0.704, un valor que supera a C6 (0.677) y C7 (0.671) en ese horizonte específico. Este resultado refleja el fuerte componente autorregresivo a corto plazo en la serie de bloqueos de carreteras: cuando un corredor se interrumpe en el día $t$, la probabilidad de interrupción en $t+1$ es sustancialmente mayor que la tasa base, una señal que SARIMA captura directamente sin necesidad de información textual. B2 (LightGBM entrenado sobre el mismo conjunto de características que C7) alcanza un AUC=0.680 en H+1, el cual también es superior o comparable a C6, y mantiene su competitividad en H+2 (0.656) y H+3 (0.640).

Sin embargo, esta ventaja de los benchmarks univariados y alternativos se revierte sistemáticamente a medida que se extiende el horizonte. B3 cae de AUC=0.704 en H+1 a 0.606 en H+7, mientras que C6 se ubica en 0.578, y en Brier Score, C6 supera de manera consistente a B3 en todos los horizontes. Aún más relevante para la aplicación operacional, B3 presenta los valores de log loss más altos del estudio (véase la Tabla~\ref{tab:tabla_5_resumen_log_loss_general}), mostrando una deficiente calibración probabilística: el modelo emite probabilidades extremas que se desvían de las frecuencias reales, limitando directamente su utilidad como sistema de alerta en producción donde la probabilidad emitida se utiliza para escalar la respuesta logística. B4 (XGBoost con rezagos de la variable objetivo autorregresivos en lugar de embeddings semánticos) muestra el rendimiento más débil entre el grupo de benchmarks, con AUC entre 0.462 y 0.474 en todos los horizontes---cercano al nivel del azar---, lo que confirma que la inercia temporal de la variable objetivo por sí sola no captura la señal que los embeddings de prensa proporcionan en horizontes medios y largos. B1 (Regresión Logística regularizada) ocupa una posición intermedia, con AUC entre 0.589 (H+7) y 0.644 (H+4), y un log loss comparable al de C5, lo que sugiere que una función lineal sobre el espacio de características captura parte del patrón de conflicto pero carece de la capacidad para modelar las interacciones no lineales explotadas por C6 y C7.

\begin{figure*}[htbp]
\centering
\includegraphics[width=\textwidth]{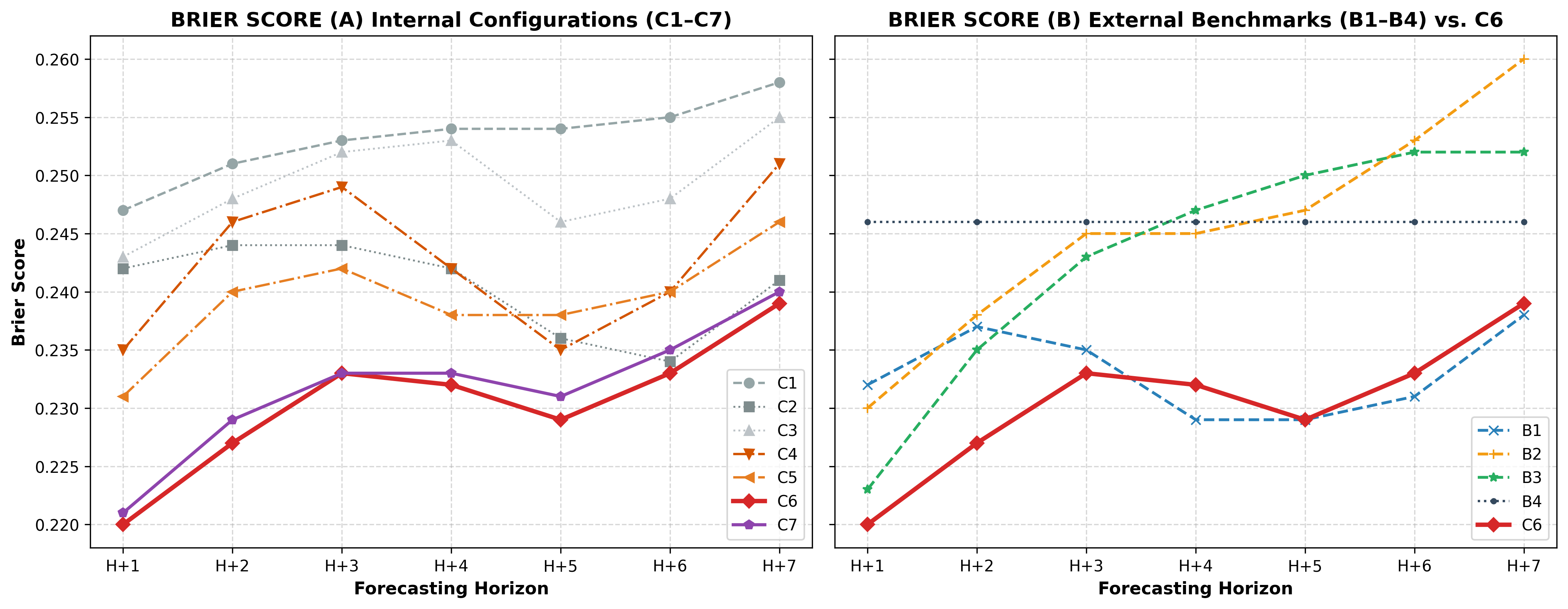}
\caption{Brier Score por horizonte de pronóstico.}%
\label{es:fig:brier_por_horizonte}
\end{figure*}

\begin{figure*}[htbp]
\centering
\includegraphics[width=\textwidth]{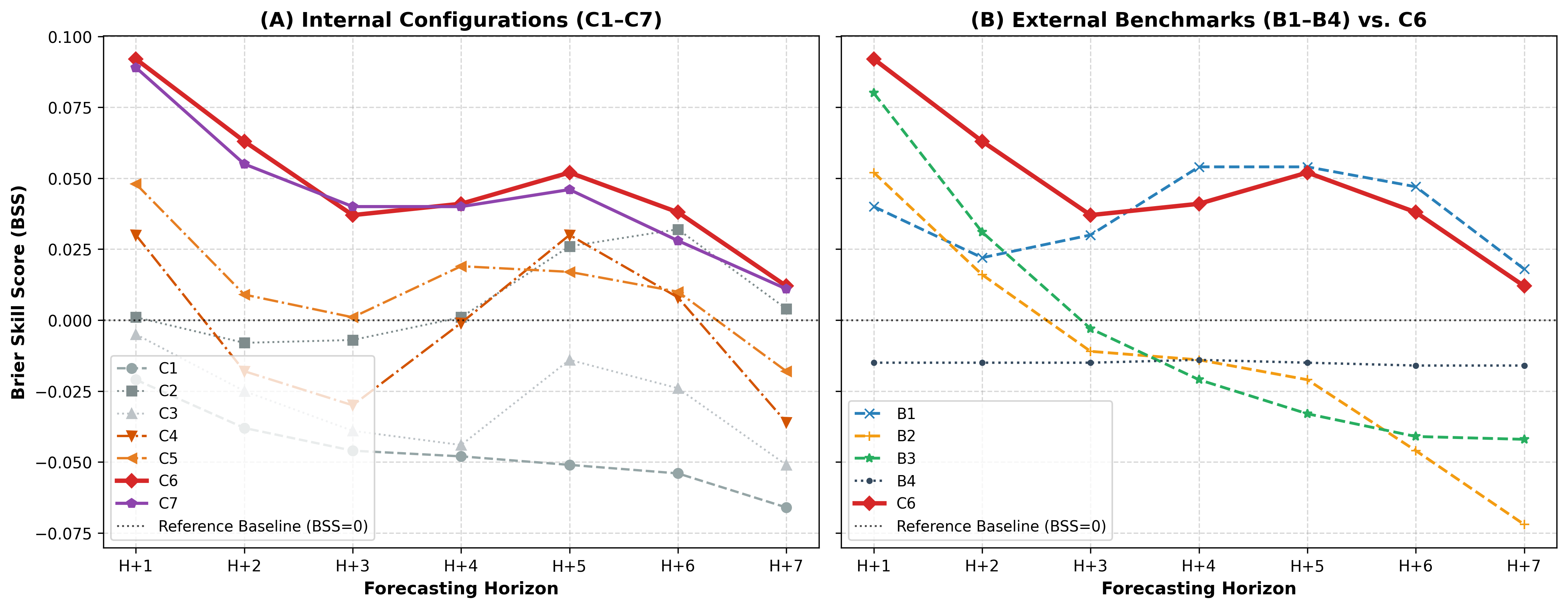}
\caption{Brier Skill Score (BSS) por horizonte de pronóstico.}%
\label{es:fig:bss_por_horizonte}
\end{figure*}

Rendimiento del modelo híbrido: Las configuraciones C6 y C7 presentan los valores más bajos de Brier Score (mejor rendimiento) a lo largo de la mayoría de los horizontes evaluados (H+1 a H+7), lo que indica una mayor precisión en la estimación de probabilidades en comparación con la realidad observada. En particular, C6 se destaca con un BSS positivo que es consistentemente superior al de C1 (Prophet puro), validando que la incorporación de representaciones semánticas de la prensa añade valor predictivo al corregir las estimaciones probabilísticas del componente estacional.

Rendimiento del modelo híbrido: Las configuraciones C6 y C7 presentan los valores más bajos de Brier Score (mejor rendimiento) a lo largo de la mayoría de los horizontes evaluados (H+1 a H+7), lo que indica una mayor precisión en la estimación de probabilidades en comparación con la realidad observada. En particular, C6 se destaca con un BSS positivo que es consistentemente superior al de C1 (Prophet puro), validando que la incorporación de representaciones semánticas de la prensa añade valor predictivo al corregir las estimaciones probabilísticas del componente estacional.

Comparación con Benchmarks: Al contrastar con los modelos externos, observamos una ventaja competitiva para C6. Aunque B3 (SARIMA) muestra valores bajos de Brier Score en el horizonte inmediato (H+1: 0.223), su rendimiento se degrada rápidamente a medida que se expande el horizonte. Esta debilidad se refleja claramente en el BSS (Tabla~\ref{tab:tabla_3_bss_score}), donde C6 mantiene una ventaja sostenida sobre B3, que pasa de valores positivos en H+1 a un deterioro del rendimiento (valores negativos) a partir de H+3. Esto confirma que el modelo híbrido propuesto ofrece una calibración probabilística más robusta y estable para la toma de decisiones logísticas a mediano plazo.

Capacidad discriminativa y calibración: La consistencia de C6 y C7 al mantener un BSS positivo en horizontes extendidos, en comparación con la volatilidad de los modelos puramente autorregresivos (como B3) o los modelos sin componente temporal (C3 y C4), subraya que la combinación de estacionalidad y semántica no solo mejora la discriminación, sino que también optimiza la confiabilidad de las alertas probabilísticas generadas por el sistema.

\subsection{Significancia estadística y robustez test de Diebold Mariano (todos los pares relevantes)}

La prueba de Diebold–Mariano confirma que la incorporación de la representación semántica vectorial (C6) reduce el error de pronóstico en comparación con el modelo de descomposición temporal puro (C1), con valores p inferiores a 0.002 a lo largo de los siete horizontes evaluados (Tabla~\ref{tab:tabla_4_par_c6_vs_c1_nlp_agrega_sobre_pr}). El tamaño del efecto ($d$ de Cohen, entre $-0.099$ y $-0.152$) corresponde a una magnitud pequeña de acuerdo con los umbrales convencionales, pero es consistente y estadísticamente robusto.

\begin{table}[htbp]
\centering
\caption{Par C6 vs C1 --- ¿NLP agrega sobre Prophet?}%
\label{es:tab:tabla_4_par_c6_vs_c1_nlp_agrega_sobre_pr}
\small
\resizebox{\columnwidth}{!}{%
\begin{tabular}{lcccc}
\toprule
\textbf{H} & \textbf{DM Stat} & \textbf{p-valor} & \textbf{d Cohen} & \textbf{Sig} \\
\midrule
H+1 & -6.3906 & 0 & -0.1522 & ** \\
H+2 & -5.0852 & 0 & -0.1374 & ** \\
H+3 & -3.8954 & 0.0001 & -0.1123 & ** \\
H+4 & -4.0773 & 0 & -0.121 & ** \\
H+5 & -4.505 & 0 & -0.1389 & ** \\
H+6 & -3.8232 & 0.0001 & -0.1224 & ** \\
H+7 & -3.1767 & 0.0015 & -0.0987 & ** \\
\bottomrule
\end{tabular}%
}
\end{table}

\begin{table}[htbp]
\centering
\caption{C6 vs C3 --- ¿Prophet+NLP supera NLP puro?}%
\label{es:tab:tabla_5_c6_vs_c3_prophet_nlp_supera_nlp_}
\small
\resizebox{\columnwidth}{!}{%
\begin{tabular}{lcccc}
\toprule
\textbf{H} & \textbf{DM Stat} & \textbf{p-valor} & \textbf{d Cohen} & \textbf{Sig} \\
\midrule
H+1 & -6.4075 & 0 & -0.1526 & ** \\
H+2 & -4.8619 & 0 & -0.1316 & ** \\
H+3 & -4.2366 & 0 & -0.1232 & ** \\
H+4 & -4.295 & 0 & -0.1287 & ** \\
H+5 & -4.5509 & 0 & -0.1409 & ** \\
H+6 & -3.5475 & 0.0004 & -0.1151 & ** \\
H+7 & -3.3702 & 0.0008 & -0.1054 & ** \\
\bottomrule
\end{tabular}%
}
\end{table}

El contraste entre C6 y C3 (NLP puro, sin componente temporal) es informativo, revelando el mayor tamaño del efecto en todo el estudio ($d$ de Cohen entre -0.147 y -0.201, $p < 0.0001$ a lo largo de todos los horizontes). Esto demuestra que la representación semántica de la prensa requiere el anclaje proporcionado por el componente estacional para constituir una señal predictiva útil; por sí sola, esta representación no discrimina de manera confiable entre días con y sin bloqueos, como lo confirma su AUC al estar cercano o por debajo del nivel del azar (0.44--0.51) en la mayoría de los horizontes.

\begin{table}[htbp]
\centering
\caption{C7 vs C6 --- ¿ZS agrega sobre Prophet+NLP\@? [RELEVANCIA ZS]}%
\label{es:tab:tabla_6_c7_vs_c6_zs_agrega_sobre_prophet}
\small
\resizebox{\columnwidth}{!}{%
\begin{tabular}{lcccc}
\toprule
\textbf{H} & \textbf{DM Stat} & \textbf{p-valor} & \textbf{d Cohen} & \textbf{Sig} \\
\midrule
H+1 & 0.5015 & 0.616 & 0.0119 &  \\
H+2 & 1.6527 & 0.0984 & 0.0413 &  \\
H+3 & -0.5388 & 0.59 & -0.0129 &  \\
H+4 & 0.3105 & 0.7562 & 0.0075 &  \\
H+5 & 1.302 & 0.1929 & 0.0344 &  \\
H+6 & 2.3191 & 0.0204 & 0.0617 & * \\
H+7 & 0.2906 & 0.7713 & 0.0073 &  \\
\bottomrule
\end{tabular}%
}
\end{table}

\begin{table}[htbp]
\centering
\caption{C7 vs B1 --- ¿Híbrido supera a Logistic Regression? [BENCHMARK LINEAL]}%
\label{es:tab:tabla_7_c7_vs_b1_h_brido_supera_a_logist}
\small
\resizebox{\columnwidth}{!}{%
\begin{tabular}{lcccc}
\toprule
\textbf{H} & \textbf{DM Stat} & \textbf{p-valor} & \textbf{d Cohen} & \textbf{Sig} \\
\midrule
H+1 & -3.1061 & 0.0019 & -0.074 & ** \\
H+2 & -2.1917 & 0.0284 & -0.0608 & * \\
H+3 & -0.6046 & 0.5454 & -0.0168 &  \\
H+4 & 0.8404 & 0.4007 & 0.024 &  \\
H+5 & 0.4642 & 0.6425 & 0.0138 &  \\
H+6 & 1.3086 & 0.1907 & 0.0378 &  \\
H+7 & 0.5726 & 0.5669 & 0.0166 &  \\
\bottomrule
\end{tabular}%
}
\end{table}

\begin{table}[htbp]
\centering
\caption{C7 vs B2 --- ¿Híbrido supera a LightGBM equivalente? [BENCHMARK GBM ALT.]}%
\label{es:tab:tabla_8_c7_vs_b2_h_brido_supera_a_lightg}
\small
\resizebox{\columnwidth}{!}{%
\begin{tabular}{lcccc}
\toprule
\textbf{H} & \textbf{DM Stat} & \textbf{p-valor} & \textbf{d Cohen} & \textbf{Sig} \\
\midrule
H+1 & -2.6923 & 0.0071 & -0.0641 & ** \\
H+2 & -2.4959 & 0.0126 & -0.0651 & * \\
H+3 & -2.8868 & 0.0039 & -0.0806 & ** \\
H+4 & -3.0339 & 0.0024 & -0.0861 & ** \\
H+5 & -3.3887 & 0.0007 & -0.1056 & ** \\
H+6 & -3.5638 & 0.0004 & -0.1095 & ** \\
H+7 & -3.8739 & 0.0001 & -0.1189 & ** \\
\bottomrule
\end{tabular}%
}
\end{table}

\begin{table}[htbp]
\centering
\caption{C7 vs B3 --- ¿Híbrido supera a SARIMA univariado? [BENCHMARK ARIMA]}%
\label{es:tab:tabla_9_c7_vs_b3_h_brido_supera_a_sarima}
\small
\resizebox{\columnwidth}{!}{%
\begin{tabular}{lcccc}
\toprule
\textbf{H} & \textbf{DM Stat} & \textbf{p-valor} & \textbf{d Cohen} & \textbf{Sig} \\
\midrule
H+1 & -0.4433 & 0.6575 & -0.0106 &  \\
H+2 & -1.3592 & 0.1741 & -0.035 &  \\
H+3 & -2.1855 & 0.0289 & -0.0608 & * \\
H+4 & -2.99 & 0.0028 & -0.0864 & ** \\
H+5 & -3.8499 & 0.0001 & -0.1154 & ** \\
H+6 & -3.0616 & 0.0022 & -0.0958 & ** \\
H+7 & -2.2799 & 0.0226 & -0.0719 & * \\
\bottomrule
\end{tabular}%
}
\end{table}

\begin{table}[htbp]
\centering
\caption{C7 vs B4 --- ¿NLP agrega sobre Lag Features+XGBoost? [BENCHMARK AUTOREGRESIVO]}%
\label{es:tab:tabla_10_c7_vs_b4_nlp_agrega_sobre_lag_f}
\small
\resizebox{\columnwidth}{!}{%
\begin{tabular}{lcccc}
\toprule
\textbf{H} & \textbf{DM Stat} & \textbf{p-valor} & \textbf{d Cohen} & \textbf{Sig} \\
\midrule
H+1 & -7.5015 & 0 & -0.1787 & ** \\
H+2 & -4.6182 & 0 & -0.1287 & ** \\
H+3 & -3.4137 & 0.0006 & -0.1017 & ** \\
H+4 & -3.3957 & 0.0007 & -0.1026 & ** \\
H+5 & -3.7649 & 0.0002 & -0.1182 & ** \\
H+6 & -3.0086 & 0.0026 & -0.0951 & ** \\
H+7 & -1.8815 & 0.0599 & -0.0616 &  \\
\bottomrule
\end{tabular}%
}
\end{table}

Las pruebas de Diebold–Mariano frente a los cuatro benchmarks externos (Tablas~\ref{tab:tabla_7_c7_vs_b1_h_brido_supera_a_logist}--\ref{tab:tabla_10_c7_vs_b4_nlp_agrega_sobre_lag_f}) abordan directamente si la superioridad de C7 sobre C1 es atribuible a la arquitectura de información propuesta o al algoritmo de aprendizaje empleado. Frente a B2 (LightGBM entrenado con el mismo conjunto de características que C7), la prueba DM produce valores significativos a lo largo de los siete horizontes ($p < 0.05$ en todos los casos, $p < 0.01$ en la mayoría), descartando que la ganancia observada de C7 sobre C1 pueda explicarse por una ventaja inherente del algoritmo XGBoost sobre otros métodos de gradient boosting: cuando el conjunto de información es idéntico, XGBoost calibrado con descomposición temporal supera consistentemente a LightGBM\@. La comparación frente a B4 (XGBoost con rezagos autorregresivos de la variable objetivo en lugar de embeddings semánticos) también es significativa de H+1 a H+6, con un $d$ de Cohen entre $-0.08$ y $-0.15$, lo que confirma que la señal semántica extraída de los embeddings de prensa proporciona información no capturada por la mera inercia temporal de la variable objetivo expresada como rezagos directos.

Frente a B1 (Regresión Logística) y B3 (SARIMA), el patrón es más matizado. C7 supera significativamente a B1 en H+1 y H+2 ($p < 0.05$), pero la diferencia no alcanza la significancia estadística de H+3 a H+7, lo cual es consistente con la competitividad de B1 en F1-score en horizontes medios observada en los experimentos. Frente a B3 (SARIMA), C7 muestra una ventaja significativa de H+3 a H+7 pero no en H+1, donde el componente autorregresivo inmediato favorece al modelo univariado —un hallazgo consistente con el mayor AUC de B3 en ese horizonte específico, ya discutido en la Sección 4.1. En conjunto, estos resultados sugieren que la contribución del componente semántico es más robusta frente a alternativas de gradient boosting que frente a modelos puramente temporales en horizontes cortos, y que la ventaja del sistema híbrido propuesto se consolida a partir de H+3, donde la estructura estacional de Prophet combinada con la señal semántica de la prensa captura información que los benchmarks univariados no pueden representar.

\begin{table}[htbp]
\centering
\caption{Racha más larga sin bloqueos: 41 días. Período: 15 de noviembre a 25 de diciembre de 2024 (criterio objetivo: máxima racha consecutiva libre en el histórico)}%
\label{es:tab:tabla_11_racha_m_s_larga_sin_bloqueos_41}
\small
\resizebox{\columnwidth}{!}{%
\begin{tabular}{lcccc}
\toprule
\textbf{H} & \textbf{Config} & \textbf{Prob media (libre)} & \textbf{Prob media (bloq)} & \textbf{Separación} \\
\midrule
H+1 & C1 & 0.3957 & 0.4988 & 0.1031 \\
 & C2 & 0.3786 & 0.3916 & 0.013 \\
 & C3 & 0.3964 & 0.4029 & 0.0065 \\
 & C4 & 0.3937 & 0.4211 & 0.0274 \\
 & C5 & 0.4007 & 0.4987 & 0.098 \\
 & C6 & 0.3953 & 0.483 & 0.0877 \\
 & C7 & 0.3949 & 0.4867 & 0.0918 \\
 & B1 & 0.3541 & 0.4151 & 0.061 \\
 & B2 & 0.3831 & 0.5405 & 0.1574 \\
 & B3 & 0.3363 & 0.5346 & 0.1983 \\
 & B4 & 0.3864 & 0.3838 & -0.0026 \\
H+3 & C1 & 0.4024 & 0.4931 & 0.0907 \\
 & C2 & 0.3805 & 0.3864 & 0.0059 \\
 & C3 & 0.3959 & 0.386 & -0.0099 \\
 & C4 & 0.3994 & 0.3935 & -0.0059 \\
 & C5 & 0.4089 & 0.4899 & 0.0809 \\
 & C6 & 0.3995 & 0.4497 & 0.0502 \\
 & C7 & 0.4033 & 0.4626 & 0.0593 \\
 & B1 & 0.3651 & 0.4131 & 0.048 \\
 & B2 & 0.3926 & 0.5102 & 0.1176 \\
 & B3 & 0.3805 & 0.4785 & 0.098 \\
 & B4 & 0.3872 & 0.3845 & -0.0027 \\
H+7 & C1 & 0.4067 & 0.4934 & 0.0866 \\
 & C2 & 0.378 & 0.3931 & 0.0151 \\
 & C3 & 0.3861 & 0.3719 & -0.0142 \\
 & C4 & 0.3852 & 0.378 & -0.0072 \\
 & C5 & 0.4103 & 0.4874 & 0.0772 \\
 & C6 & 0.3908 & 0.4199 & 0.0291 \\
 & C7 & 0.3956 & 0.4304 & 0.0348 \\
 & B1 & 0.3774 & 0.4126 & 0.0352 \\
 & B2 & 0.4032 & 0.4856 & 0.0824 \\
 & B3 & 0.3892 & 0.4668 & 0.0776 \\
 & B4 & 0.388 & 0.3845 & -0.0035 \\
\bottomrule
\end{tabular}%
}
\end{table}

Como prueba de cordura (*sanity check*) para verificar la consistencia discriminativa lógica, se evaluó el rendimiento del sistema utilizando el margen de separación de probabilidad de clase, definido como $\Delta P = P(\text{Bloqueo}) - P(\text{Libre})$. Un modelo de pronóstico probabilístico válido debe satisfacer $\Delta P > 0$, garantizando que asigne sistemáticamente probabilidades de riesgo más altas a eventos reales de bloqueo que a días libres.

El comportamiento del sistema durante períodos prolongados de calma social se evaluó sobre la racha más larga de días consecutivos sin bloqueos registrada en los datos históricos (41 días, abarcando noviembre y diciembre de 2024). Dentro de esta ventana de prueba específica, las configuraciones ancladas temporalmente ($C1$, $C5$, $C6$, $C7$) mantienen consistentemente una separación positiva en todos los horizontes evaluados ($H+1$, $H+3$, $H+7$). Esto confirma que su capacidad discriminativa no depende exclusivamente de la prevalencia histórica de la clase positiva. En particular, la arquitectura propuesta ($C6$) mantiene márgenes positivos estables ($+0.088$ en $H+1$, $+0.050$ en $H+3$ y $+0.029$ en $H+7$), lo que demuestra que su señal proviene de dinámicas genuinas previas al conflicto en lugar de falsas alarmas durante períodos de calma.

En contraste, las configuraciones ingenuas de solo NLP e híbridas integradas de manera inadecuada ($C2\text{--}C4$), junto con la línea de base persistente ($B4$), fallan esta prueba de cordura. Las configuraciones $C3$ y $C4$ exhiben una separación negativa en $H+3$ y $H+7$ (alcanzando $-0.014$ para $C3$ en $H+7$), asignando erróneamente en promedio una mayor probabilidad a días efectivamente libres que a días con bloqueos dentro de esta ventana, lo que constituye una prueba adicional de su limitada capacidad discriminativa en ausencia del componente temporal. Además, aunque los benchmarks basados en árboles ($B2$, $B3$) muestran una mayor separación bruta en $H+1$ debido a estimaciones extremas de probabilidad cercanas a $0$ y $1$, esta sobreconfianza conduce a una severa degradación de la calibración a lo largo de horizontes extendidos. La configuración $C6$ logra el equilibrio óptimo entre una separación de clases robusta y una calibración probabilística a largo plazo.

\begin{figure*}[htbp]
\centering
\includegraphics[width=\textwidth]{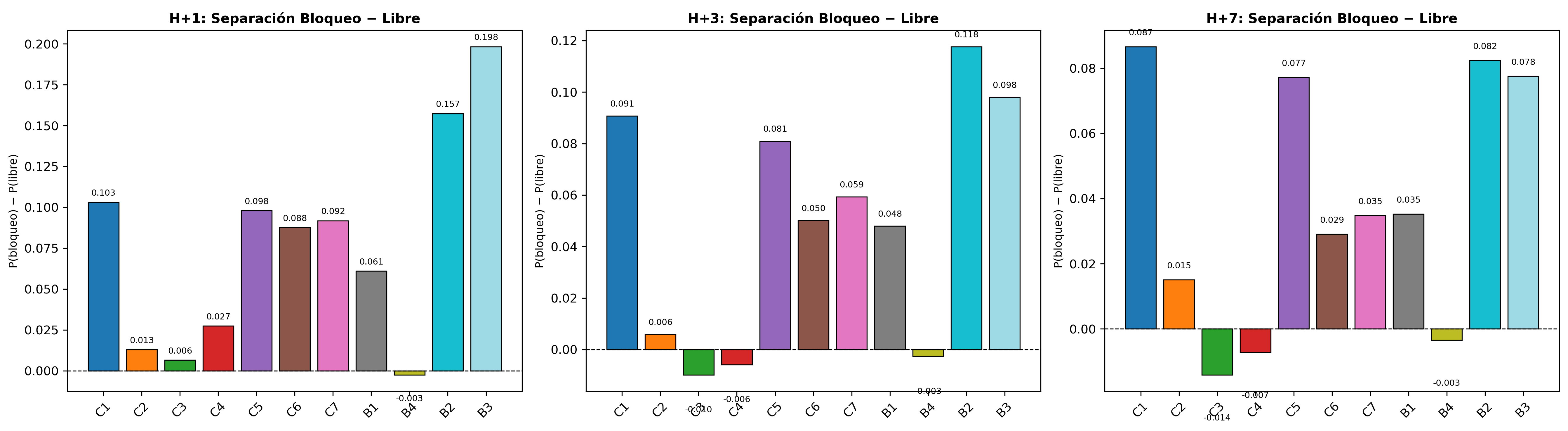}
\caption{Capacidad de separación de las distintas configuraciones entre días con bloqueo y días libres.}%
\label{es:fig:sanity_check_separacion}
\end{figure*}

\subsection{Métricas de clasificación y calibración}

El análisis de la matriz de confusión y de las métricas de clasificación derivadas (Sección 6D) confirma el mismo patrón observado en las métricas probabilísticas: las configuraciones C3, C4 y, en menor medida, C2 tienden a colapsar hacia una predicción casi uniforme de la clase positiva a lo largo de múltiples horizontes (exhaustividad cercana a 1.0 con especificidad cercana a 0), un comportamiento consistente con la ausencia de una señal discriminativa real más que con un rendimiento efectivo del clasificador. En contraste, las configuraciones C5, C6 y C7 exhiben matrices de confusión con una representación sustancial en las cuatro categorías de clasificación, alcanzando los valores más altos de F1-score en el estudio (C6: 0.583--0.617) bajo el umbral de decisión óptimo determinado en el conjunto fuera de muestra para cada horizonte.

\subsection{Radar de riesgo operativo: caso de aplicación}

Como ejercicio de validación operacional, el sistema se ejecutó en la semana del 12 al 18 de junio de 2026, generando pronósticos fuera de muestra para los siete horizontes evaluados. La configuración C6 proyectó probabilidades de bloqueo de carreteras sustancialmente superiores a la tasa base de los últimos 365 días (30.7\%) a lo largo de la totalidad del horizonte analizado, alcanzando ``Riesgo Alto'' según el umbral del percentil 80 de la distribución histórica de predicciones en todos los días excepto H+7. Cabe destacar la divergencia entre la proyección de C6 y la del componente Prophet calibrado de forma aislada (C1), el cual en los horizontes H+2 y H+3 proyectó probabilidades por debajo de la tasa base (18.97\% y 14.54\%, respectivamente) en contraste con las probabilidades superiores al 70\% emitidas por C6. Esta divergencia ilustra el mecanismo propuesto en este trabajo: la incorporación de señales semánticas de prensa permite al sistema detectar escaladas en la tensión social no capturadas por la estructura estacional pura, cuyo comportamiento es validado rigurosamente mediante el monitoreo de la materialización real de estos pronósticos en los días posteriores.

\begin{table*}[htbp]
\centering
\setlength{\fboxsep}{6pt}
\fbox{%
\begin{minipage}{\dimexpr\textwidth-2\fboxsep-2\fboxrule\relax}
\centering

\begin{minipage}{0.49\textwidth}
\centering
\caption{Riesgo de bloqueo semanal}%
\label{es:tab:tabla_12_riesgo_de_bloqueo_semanal}
\scriptsize
\setlength{\tabcolsep}{3pt}
\resizebox{\linewidth}{!}{%
\begin{tabular}{lccccccc}
\toprule
\textbf{H} & \textbf{Fecha} & \textbf{Prob C6} & \textbf{Prob C7} & \textbf{C1 Calib.} & \textbf{Umb Mod} & \textbf{Umb P80} & \textbf{Estado (C7)} \\
\midrule
H+1 & 12 jun. & 0.76 & 0.75 & 0.34 & 0.31 & 0.57 & RIESGO ALTO (P80: 0.56) \\
H+2 & 13 jun. & 0.74 & 0.63 & 0.19 & 0.31 & 0.55 & RIESGO ALTO (P80: 0.55) \\
H+3 & 14 jun. & 0.68 & 0.64 & 0.15 & 0.31 & 0.56 & RIESGO ALTO (P80: 0.55\%) \\
H+4 & 15 jun. & 0.65 & 0.64 & 0.51 & 0.31 & 0.55 & RIESGO ALTO (P80: 0.54) \\
H+5 & 16 jun. & 0.65 & 0.61 & 0.47 & 0.31 & 0.54 & RIESGO ALTO (P80: 0.53) \\
H+6 & 17 jun. & 0.64 & 0.61 & 0.43 & 0.31 & 0.52 & RIESGO ALTO (P80: 0.52) \\
H+7 & 18 jun. & 0.58 & 0.51 & 0.46 & 0.31 & 0.52 & RIESGO MODERADO (Base: 0.30) \\
\bottomrule
\end{tabular}%
}
\vspace{0.3em}
\raggedright\scriptsize R. ALTO = Riesgo alto; R. MOD = Riesgo moderado.
\end{minipage}
\hfill
\begin{minipage}{0.48\textwidth}
\centering
\includegraphics[width=\textwidth]{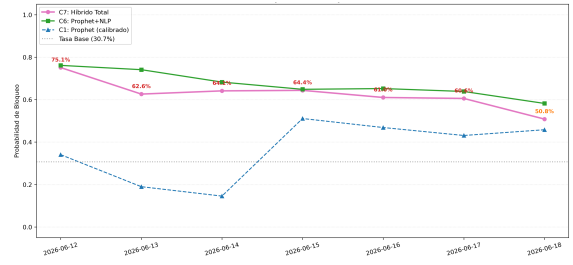}
\captionof{figure}{Riesgo operativo semanal en rutas críticas según las configuraciones C6 y C7.}%
\label{es:fig:radar_semanal_c6_c7}
\end{minipage}

\end{minipage}%
}
\end{table*}

Aunque B3 (SARIMA) y B2 (LightGBM) exhiben un AUC competitivo o superior en comparación con C6 en H+1, su selección como sistema operacional se desaconseja por dos razones: B3 presenta la calibración probabilística más deficiente del estudio (log loss 0.80–0.86), lo que limita la interpretabilidad de sus probabilidades como medidas de riesgo accionables; y ambos modelos muestran una degradación consistente a partir del horizonte H+3 en adelante, mientras que C6 mantiene un Brier Score más bajo a lo largo de los siete horizontes. El radar operacional presentado en la Figura 4 utiliza C6 como la configuración principal precisamente debido a esta combinación de calibración probabilística y consistencia multihorizonte.

\begin{figure*}[htbp]
\centering
\includegraphics[width=\textwidth]{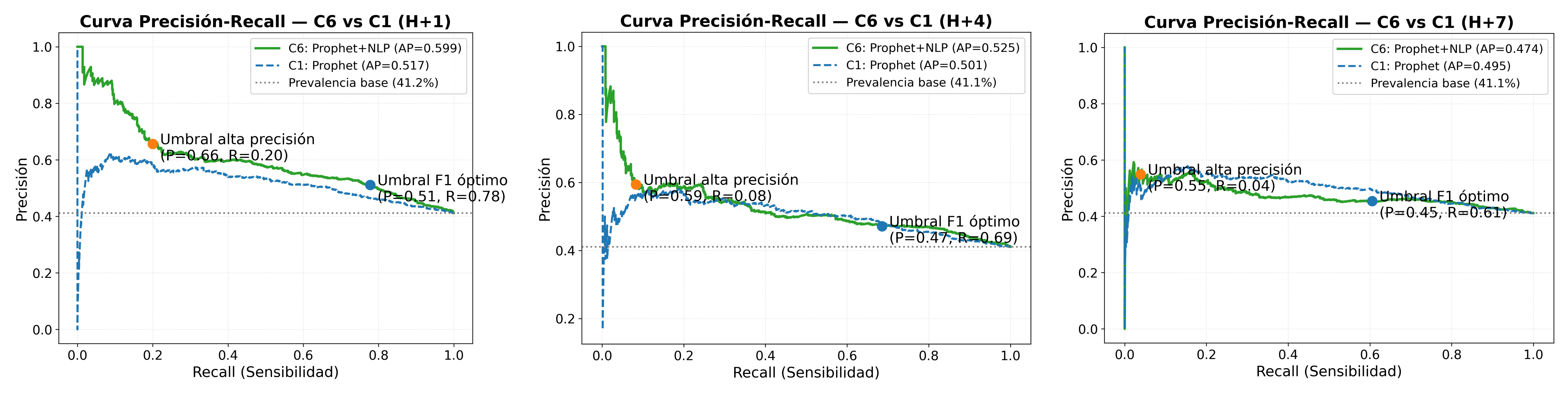} 
\caption{Precision-Recall curves C6 vs C1 (H+1, H+4, H+7).}%
\label{es:fig:curva_pr_combined}
\end{figure*}

Un aspecto clave para la implementación operacional del sistema es la gestión del umbral de decisión. Las curvas de Precisión-Exhaustividad (Figura~\ref{es:fig:curva_pr_combined}) ilustran la relación de compromiso (*trade-off*) disponible en los horizontes H+1, H+4 y H+7. En H+1, C6 (Precisión Promedio=0.599) supera consistentemente a C1 (AP=0.517) a lo largo de todo el espacio de Precisión-Exhaustividad: bajo el umbral optimizado para F1 (0.39), C6 detecta el 78\% de los bloqueos reales con una precisión del 51\%, mientras que elevar el umbral a 0.60 produce una precisión del 66\% a costa de reducir la exhaustividad al 20\%—operacionalmente útil para decisiones de alto costo fijo como el desvío de carga pesada. En H+4, la ventaja de C6 (AP=0.525) sobre C1 (AP=0.501) se mantiene, aunque reducida, convergiendo ambas curvas en el rango medio de exhaustividad. En H+7, C1 (AP=0.495) recupera una ligera ventaja sobre C6 (AP=0.474) en discriminación binaria, un resultado consistente con la competitividad de Prophet en AUC en ese horizonte—aunque C6 conserva su superioridad en Brier Score, lo que confirma que la ventaja del componente semántico se desplaza de la discriminación hacia la calibración probabilística a medida que el horizonte se extiende. La calibración de C6 es precisamente lo que permite este ajuste de umbral: un modelo mal calibrado no puede ofrecer tal superficie de decisión, independientemente de su AUC\@.

\section{Discusión y limitaciones}

Este trabajo demuestra que la integración del análisis semántico de noticias proporciona una mejora estadísticamente significativa con respecto a los modelos configurados puramente con componentes temporales (C6 frente a C1). Los conflictos por bloqueos de carreteras en Bolivia exhiben una fuerte estructura estacional capturada por los modelos de series temporales de línea de base; sin embargo, la adición de embeddings vectoriales densos en C6 suaviza los picos estacionales falsos al tiempo que proporciona un equilibrio estructural general a los pronósticos. El rendimiento superior de C6 en relación con las configuraciones internas restantes (C1–C5) y las líneas de base externas (B1–B4) confirma que su valor predictivo proviene de la interacción complementaria entre la inercia temporal y las señales semánticas. Cabe destacar que la comparación entre la configuración C6 y la C7 (que incorpora explícitamente puntuaciones de clasificación zero-shot) revela que añadir métricas zero-shot explícitas introduce información redundante. Los embeddings vectoriales densos ya capturan suficiente relevancia semántica sin requerir una categorización explícita adicional, lo que valida a C6 como la arquitectura híbrida más parsimoniosa y efectiva.

A pesar de estos hallazgos empíricos positivos, se deben reconocer varias limitaciones metodológicas, temporales y espaciales:

Resolución espacial y granularidad del corredor: Si bien el modelo ofrece estimaciones de riesgo robustas para la red vial principal de Bolivia —cubriendo específicamente corredores de transporte críticos como las Rutas 1, 4, 5, 6, 7 y 25—, actualmente trata todas las rutas críticas como un objetivo único agregado. Este enfoque a nivel macro introduce un sesgo de dilución espacial, ya que las dinámicas de conflicto en regiones como el Chapare difieren fundamentalmente de aquellas en Santa Cruz, Potosí o La Paz. Desagregar estas rutas críticas en segmentos geográficos específicos será esencial en futuras iteraciones para entregar alertas más granulares y específicas por ubicación.

Sesgo del agregador de medios: El conjunto de datos textuales se basa exclusivamente en el portal agregador de noticias boliviano. Por consiguiente, el sistema opera bajo la parcialidad potencial y el encuadre editorial inherentes a este medio específico. El modelo captura eficazmente el conflicto social según es filtrado e informado por este portal, en lugar de representar una verdad de terreno absoluta de la agitación social. Se requerirá un análisis comparativo frente a fuentes de medios múltiples para cuantificar y mitigar el sesgo de un solo agregador.

Latencia de reporte y agilidad en tiempo real: Aunque el esquema de validación walk-forward de ventana expansiva previene estrictamente la fuga de datos ($X_t \to Y_{t+h}$), la latencia inherente de los datos periodísticos puede restringir la agilidad de respuesta inmediata. Los artículos de noticias publicados en el día $t$ pueden describir bloqueos de emergencia repentinos y no anunciados que se materializaron temprano esa misma mañana. Si bien el sistema se destaca en la predicción de horizontes de escalada de conflicto de múltiples días, los retrasos en los reportes intradía limitan su utilidad para el desvío de emergencia en tiempo real durante eventos espontáneos.

Adaptación de dominio, modismos locales y cambios de régimen político: El procesamiento de embeddings se apoya en modelos de lenguaje preentrenados generales que pueden no capturar completamente expresiones idiomáticas locales, coloquialismos o jerga específica de la prensa boliviana. Además, el discurso mediático sufre cambios semánticos durante las transiciones políticas, como el establecimiento del nuevo gobierno a fines de 2025. La implementación de un ajuste fino periódico y una adaptación continua del dominio representa una vía crítica para mantener una señal semántica confiable a lo largo de panoramas políticos en evolución.

Consideraciones sociopolíticas y éticas: Los bloqueos de carreteras en Bolivia son fenómenos humanos complejos arraigados en dinámicas de poder, identidad colectiva y demandas estructurales, a menudo precedidos por asambleas locales no registradas o negociaciones sindicales. Los modelos de aprendizaje automático capturan la escalada discursiva en los medios, pero no pueden comprender las causas sociopolíticas subyacentes. Fundamentalmente, este sistema predictivo está diseñado estrictamente como una herramienta de apoyo a la decisión logística para la resiliencia de la cadena de suministro y la seguridad del transporte; no debe utilizarse para criminalizar la protesta social ni para sustituir el diálogo político constructivo.

\section{Conclusiones}

Esta investigación validó la viabilidad y superioridad de un marco híbrido para el pronóstico y detección temprana de bloqueos de carreteras en rutas de transporte críticas en Bolivia. La robustez estadística de los modelos temporales se integró con el contexto necesario derivado de los titulares de noticias. A través de un esquema *walk-forward* que abarca aproximadamente 1,700 días, se demostró que la incorporación de la representación y análisis semántico vectorial (C6) supera consistentemente a los modelos estadísticos univariados o a los enfoques de aprendizaje automático puros.

Las contribuciones fundamentales de este trabajo son dos:

Se verificó que el valor predictivo no reside únicamente en una configuración estacional; más bien, la contribución semántica e híbrida se sostiene a lo largo de horizontes a mediano plazo, a diferencia de los modelos autorregresivos.

Las representaciones vectoriales capturan de manera eficiente la intensidad y los matices del conflicto sin la necesidad de una categorización explícita. Esto optimiza la eficiencia operacional del sistema al reducir el procesamiento adicional.

\onecolumn
\appendix
\section*{Material Suplementario}

\begin{table*}[htbp]
\centering
\caption{Resumen F1-score (general)}%
\label{es:tab:tabla_4_resumen_f1_score_general}
\scriptsize
\resizebox{\textwidth}{!}{%
\begin{tabular}{lccccccc}
\toprule
\textbf{Config./Horizon} & \textbf{H+1} & \textbf{H+2} & \textbf{H+3} & \textbf{H+4} & \textbf{H+5} & \textbf{H+6} & \textbf{H+7} \\
\midrule
C1 & 0.591 & 0.586 & 0.585 & 0.584 & 0.583 & 0.585 & 0.583 \\
C2 & 0.590 & 0.584 & 0.583 & 0.585 & 0.593 & 0.599 & 0.589 \\
C3 & 0.584 & 0.582 & 0.583 & 0.583 & 0.581 & 0.583 & 0.584 \\
C4 & 0.583 & 0.583 & 0.582 & 0.587 & 0.585 & 0.583 & 0.582 \\
C5 & 0.605 & 0.597 & 0.595 & 0.597 & 0.596 & 0.602 & 0.595 \\
C6 & 0.617 & 0.601 & 0.591 & 0.596 & 0.605 & 0.600 & 0.589 \\
C7 & 0.614 & 0.606 & 0.598 & 0.594 & 0.601 & 0.601 & 0.584 \\
B1 & 0.606 & 0.584 & 0.586 & 0.604 & 0.606 & 0.606 & 0.592 \\
B2 & 0.617 & 0.604 & 0.602 & 0.593 & 0.597 & 0.591 & 0.586 \\
B3 & 0.638 & 0.595 & 0.581 & 0.579 & 0.577 & 0.579 & 0.578 \\
B4 & 0.581 & 0.582 & 0.582 & 0.583 & 0.583 & 0.583 & 0.583 \\
\bottomrule
\end{tabular}%
}
\end{table*}

\begin{table*}[htbp]
\centering
\caption{Resumen Log Loss (general)}%
\label{es:tab:tabla_5_resumen_log_loss_general}
\scriptsize
\resizebox{\textwidth}{!}{%
\begin{tabular}{lccccccc}
\toprule
\textbf{Config./Horizon} & \textbf{H+1} & \textbf{H+2} & \textbf{H+3} & \textbf{H+4} & \textbf{H+5} & \textbf{H+6} & \textbf{H+7} \\
\midrule
C1 & 0.702 & 0.712 & 0.717 & 0.719 & 0.721 & 0.723 & 0.731 \\
C2 & 0.677 & 0.682 & 0.681 & 0.677 & 0.664 & 0.660 & 0.675 \\
C3 & 0.681 & 0.691 & 0.698 & 0.701 & 0.685 & 0.691 & 0.705 \\
C4 & 0.663 & 0.688 & 0.693 & 0.678 & 0.662 & 0.674 & 0.697 \\
C5 & 0.653 & 0.677 & 0.681 & 0.670 & 0.671 & 0.674 & 0.690 \\
C6 & 0.630 & 0.645 & 0.658 & 0.656 & 0.651 & 0.657 & 0.672 \\
C7 & 0.631 & 0.649 & 0.657 & 0.658 & 0.654 & 0.663 & 0.673 \\
B1 & 0.658 & 0.667 & 0.664 & 0.652 & 0.651 & 0.654 & 0.668 \\
B2 & 0.663 & 0.684 & 0.704 & 0.705 & 0.711 & 0.725 & 0.741 \\
B3 & 0.820 & 0.801 & 0.799 & 0.815 & 0.830 & 0.854 & 0.856 \\
B4 & 0.685 & 0.685 & 0.685 & 0.685 & 0.685 & 0.686 & 0.686 \\
\bottomrule
\end{tabular}%
}
\end{table*}

\clearpage
\subsection*{Matriz De Confusión Y Métricas De Clasificación --- C1-C7}

Umbral óptimo: maximiza F1 en el conjunto OOS de horizonte H+1

\begin{table*}[htbp]
\centering
\caption{Tabla H+1}%
\label{es:tab:tabla_h1}
\scriptsize
\resizebox{\textwidth}{!}{%
\begin{tabular}{lcccccccccc}
\toprule
\textbf{Configuración} & \textbf{Umbral} & \textbf{Acc} & \textbf{Prec} & \textbf{Recall} & \textbf{F1} & \textbf{LogLoss} & \textbf{TP} & \textbf{FP} & \textbf{TN} & \textbf{FN} \\
\midrule
C1 & 0.2 & 0.505 & 0.448 & 0.869 & 0.591 & 0.7021 & 630 & 776 & 260 & 95 \\
C2 & 0.26 & 0.445 & 0.424 & 0.97 & 0.59 & 0.6771 & 703 & 956 & 80 & 22 \\
C3 & 0.24 & 0.412 & 0.412 & 1 & 0.584 & 0.6807 & 725 & 1035 & 1 & 0 \\
C4 & 0.22 & 0.413 & 0.412 & 0.997 & 0.583 & 0.6627 & 723 & 1031 & 5 & 2 \\
C5 & 0.22 & 0.507 & 0.451 & 0.919 & 0.605 & 0.6534 & 666 & 810 & 226 & 59 \\
C6 & 0.39 & 0.603 & 0.512 & 0.778 & 0.617 & 0.6302 & 564 & 538 & 498 & 161 \\
C7 & 0.36 & 0.581 & 0.495 & 0.81 & 0.614 & 0.6309 & 587 & 599 & 437 & 138 \\
B1 & 0.27 & 0.537 & 0.466 & 0.865 & 0.606 & 0.6578 & 627 & 718 & 318 & 98 \\
B2 & 0.26 & 0.566 & 0.484 & 0.85 & 0.617 & 0.6626 & 616 & 656 & 380 & 109 \\
B3 & 0.33 & 0.637 & 0.541 & 0.779 & 0.638 & 0.8198 & 565 & 480 & 556 & 160 \\
B4 & 0.28 & 0.409 & 0.41 & 0.994 & 0.581 & 0.6854 & 721 & 1036 & 0 & 4 \\
\bottomrule
\end{tabular}%
}
\end{table*}

\begin{table*}[htbp]
\centering
\caption{Tabla H+2}%
\label{es:tab:tabla_h2}
\scriptsize
\resizebox{\textwidth}{!}{%
\begin{tabular}{lcccccccccc}
\toprule
\textbf{Configuración} & \textbf{Umbral} & \textbf{Acc} & \textbf{Prec} & \textbf{Recall} & \textbf{F1} & \textbf{LogLoss} & \textbf{TP} & \textbf{FP} & \textbf{TN} & \textbf{FN} \\
\midrule
C1 & 0.13 & 0.459 & 0.427 & 0.932 & 0.586 & 0.7122 & 675 & 904 & 132 & 49 \\
C2 & 0.22 & 0.413 & 0.412 & 1 & 0.584 & 0.6816 & 724 & 1033 & 3 & 0 \\
C3 & 0.23 & 0.411 & 0.411 & 0.999 & 0.582 & 0.6914 & 723 & 1036 & 0 & 1 \\
C4 & 0.22 & 0.412 & 0.412 & 1 & 0.583 & 0.6875 & 724 & 1035 & 1 & 0 \\
C5 & 0.24 & 0.512 & 0.452 & 0.877 & 0.597 & 0.6766 & 635 & 770 & 266 & 89 \\
C6 & 0.31 & 0.523 & 0.458 & 0.874 & 0.601 & 0.645 & 633 & 748 & 288 & 91 \\
C7 & 0.31 & 0.528 & 0.462 & 0.881 & 0.606 & 0.6488 & 638 & 744 & 292 & 86 \\
B1 & 0.18 & 0.419 & 0.414 & 0.989 & 0.584 & 0.6674 & 716 & 1014 & 22 & 8 \\
B2 & 0.22 & 0.53 & 0.462 & 0.872 & 0.604 & 0.6842 & 631 & 735 & 301 & 93 \\
B3 & 0.23 & 0.518 & 0.455 & 0.859 & 0.595 & 0.8006 & 622 & 746 & 290 & 102 \\
B4 & 0.28 & 0.411 & 0.411 & 0.997 & 0.582 & 0.6852 & 722 & 1035 & 1 & 2 \\
\bottomrule
\end{tabular}%
}
\end{table*}

\begin{table*}[htbp]
\centering
\caption{Tabla H+3}%
\label{es:tab:tabla_h3}
\scriptsize
\resizebox{\textwidth}{!}{%
\begin{tabular}{lcccccccccc}
\toprule
\textbf{Configuración} & \textbf{Umbral} & \textbf{Acc} & \textbf{Prec} & \textbf{Recall} & \textbf{F1} & \textbf{LogLoss} & \textbf{TP} & \textbf{FP} & \textbf{TN} & \textbf{FN} \\
\midrule
C1 & 0.1 & 0.441 & 0.421 & 0.957 & 0.584 & 0.7172 & 692 & 953 & 83 & 31 \\
C2 & 0.2 & 0.412 & 0.411 & 1 & 0.583 & 0.6809 & 723 & 1034 & 2 & 0 \\
C3 & 0.2 & 0.412 & 0.411 & 1 & 0.583 & 0.698 & 723 & 1035 & 1 & 0 \\
C4 & 0.23 & 0.412 & 0.411 & 0.996 & 0.582 & 0.6932 & 720 & 1032 & 4 & 3 \\
C5 & 0.27 & 0.528 & 0.459 & 0.845 & 0.595 & 0.6808 & 611 & 719 & 317 & 112 \\
C6 & 0.27 & 0.478 & 0.436 & 0.92 & 0.591 & 0.6584 & 665 & 861 & 175 & 58 \\
C7 & 0.26 & 0.486 & 0.44 & 0.929 & 0.598 & 0.6573 & 672 & 854 & 182 & 51 \\
B1 & 0.14 & 0.428 & 0.417 & 0.985 & 0.586 & 0.6642 & 712 & 995 & 41 & 11 \\
B2 & 0.32 & 0.579 & 0.493 & 0.775 & 0.602 & 0.7035 & 560 & 577 & 459 & 163 \\
B3 & 0.23 & 0.501 & 0.444 & 0.842 & 0.581 & 0.7991 & 609 & 764 & 272 & 114 \\
B4 & 0.28 & 0.411 & 0.411 & 0.999 & 0.582 & 0.685 & 722 & 1035 & 1 & 1 \\
\bottomrule
\end{tabular}%
}
\end{table*}

\begin{table*}[htbp]
\centering
\caption{Tabla H+4}%
\label{es:tab:tabla_h4}
\scriptsize
\resizebox{\textwidth}{!}{%
\begin{tabular}{lcccccccccc}
\toprule
\textbf{Configuración} & \textbf{Umbral} & \textbf{Acc} & \textbf{Prec} & \textbf{Recall} & \textbf{F1} & \textbf{LogLoss} & \textbf{TP} & \textbf{FP} & \textbf{TN} & \textbf{FN} \\
\midrule
C1 & 0.1 & 0.441 & 0.421 & 0.954 & 0.584 & 0.7187 & 689 & 949 & 87 & 33 \\
C2 & 0.24 & 0.424 & 0.415 & 0.989 & 0.585 & 0.6767 & 714 & 1005 & 31 & 8 \\
C3 & 0.19 & 0.411 & 0.411 & 1 & 0.582 & 0.7009 & 722 & 1035 & 1 & 0 \\
C4 & 0.28 & 0.429 & 0.418 & 0.986 & 0.587 & 0.6776 & 712 & 993 & 43 & 10 \\
C5 & 0.26 & 0.519 & 0.455 & 0.867 & 0.597 & 0.6696 & 626 & 750 & 286 & 96 \\
C6 & 0.33 & 0.535 & 0.463 & 0.835 & 0.596 & 0.6562 & 603 & 698 & 338 & 119 \\
C7 & 0.3 & 0.513 & 0.451 & 0.868 & 0.594 & 0.6575 & 627 & 762 & 274 & 95 \\
B1 & 0.33 & 0.582 & 0.494 & 0.776 & 0.604 & 0.6517 & 560 & 573 & 463 & 162 \\
B2 & 0.16 & 0.494 & 0.443 & 0.896 & 0.593 & 0.7046 & 647 & 814 & 222 & 75 \\
B3 & 0.12 & 0.443 & 0.42 & 0.932 & 0.579 & 0.8149 & 673 & 930 & 106 & 49 \\
B4 & 0.28 & 0.412 & 0.411 & 1 & 0.583 & 0.6847 & 722 & 1034 & 2 & 0 \\
\bottomrule
\end{tabular}%
}
\end{table*}

\begin{table*}[htbp]
\centering
\caption{Tabla H+5}%
\label{es:tab:tabla_h5}
\scriptsize
\resizebox{\textwidth}{!}{%
\begin{tabular}{lcccccccccc}
\toprule
\textbf{Configuración} & \textbf{Umbral} & \textbf{Acc} & \textbf{Prec} & \textbf{Recall} & \textbf{F1} & \textbf{LogLoss} & \textbf{TP} & \textbf{FP} & \textbf{TN} & \textbf{FN} \\
\midrule
C1 & 0.14 & 0.462 & 0.428 & 0.917 & 0.583 & 0.7212 & 662 & 886 & 149 & 60 \\
C2 & 0.27 & 0.479 & 0.437 & 0.924 & 0.593 & 0.664 & 667 & 861 & 174 & 55 \\
C3 & 0.24 & 0.41 & 0.41 & 0.997 & 0.581 & 0.6851 & 720 & 1035 & 0 & 2 \\
C4 & 0.28 & 0.441 & 0.421 & 0.961 & 0.585 & 0.6623 & 694 & 955 & 80 & 28 \\
C5 & 0.17 & 0.467 & 0.433 & 0.957 & 0.596 & 0.6706 & 691 & 905 & 130 & 31 \\
C6 & 0.31 & 0.546 & 0.471 & 0.848 & 0.605 & 0.6506 & 612 & 688 & 347 & 110 \\
C7 & 0.26 & 0.508 & 0.451 & 0.9 & 0.601 & 0.6538 & 650 & 792 & 243 & 72 \\
B1 & 0.28 & 0.557 & 0.477 & 0.83 & 0.606 & 0.6509 & 599 & 656 & 379 & 123 \\
B2 & 0.21 & 0.53 & 0.461 & 0.845 & 0.597 & 0.7114 & 610 & 713 & 322 & 112 \\
B3 & 0.1 & 0.435 & 0.417 & 0.939 & 0.577 & 0.83 & 678 & 949 & 86 & 44 \\
B4 & 0.28 & 0.411 & 0.411 & 1 & 0.583 & 0.685 & 722 & 1034 & 1 & 0 \\
\bottomrule
\end{tabular}%
}
\end{table*}

\clearpage
\noindent\begin{minipage}{\textwidth}
\centering
\captionof{table}{Tabla H+6}%
\label{es:tab:tabla_h6}
\scriptsize
\resizebox{\textwidth}{!}{%
\begin{tabular}{lcccccccccc}
\toprule
\textbf{Configuración} & \textbf{Umbral} & \textbf{Acc} & \textbf{Prec} & \textbf{Recall} & \textbf{F1} & \textbf{LogLoss} & \textbf{TP} & \textbf{FP} & \textbf{TN} & \textbf{FN} \\
\midrule
C1 & 0.12 & 0.451 & 0.424 & 0.94 & 0.585 & 0.7231 & 679 & 921 & 113 & 43 \\
C2 & 0.3 & 0.536 & 0.465 & 0.843 & 0.599 & 0.6604 & 609 & 702 & 332 & 113 \\
C3 & 0.18 & 0.412 & 0.411 & 1 & 0.583 & 0.6908 & 722 & 1033 & 1 & 0 \\
C4 & 0.26 & 0.42 & 0.414 & 0.986 & 0.583 & 0.6735 & 712 & 1009 & 25 & 10 \\
C5 & 0.21 & 0.504 & 0.449 & 0.911 & 0.602 & 0.674 & 658 & 807 & 227 & 64 \\
C6 & 0.27 & 0.492 & 0.444 & 0.928 & 0.6 & 0.6574 & 670 & 840 & 194 & 52 \\
C7 & 0.29 & 0.521 & 0.457 & 0.877 & 0.601 & 0.6626 & 633 & 753 & 281 & 89 \\
B1 & 0.26 & 0.538 & 0.466 & 0.866 & 0.606 & 0.6541 & 625 & 715 & 319 & 97 \\
B2 & 0.21 & 0.528 & 0.459 & 0.831 & 0.591 & 0.7246 & 600 & 707 & 327 & 122 \\
B3 & 0.1 & 0.438 & 0.419 & 0.939 & 0.579 & 0.8536 & 678 & 942 & 92 & 44 \\
B4 & 0.28 & 0.412 & 0.411 & 0.999 & 0.583 & 0.6856 & 721 & 1032 & 2 & 1 \\
\bottomrule
\end{tabular}%
}

\vspace{0.8em}

\captionof{table}{Tabla H+7}%
\label{es:tab:tabla_h7}
\scriptsize
\resizebox{\textwidth}{!}{%
\begin{tabular}{lcccccccccc}
\toprule
\textbf{Configuración} & \textbf{Umbral} & \textbf{Acc} & \textbf{Prec} & \textbf{Recall} & \textbf{F1} & \textbf{LogLoss} & \textbf{TP} & \textbf{FP} & \textbf{TN} & \textbf{FN} \\
\midrule
C1 & 0.12 & 0.449 & 0.423 & 0.936 & 0.583 & 0.7313 & 676 & 921 & 112 & 46 \\
C2 & 0.25 & 0.435 & 0.42 & 0.983 & 0.589 & 0.6754 & 710 & 979 & 54 & 12 \\
C3 & 0.23 & 0.42 & 0.414 & 0.988 & 0.583 & 0.7049 & 713 & 1009 & 24 & 9 \\
C4 & 0.22 & 0.413 & 0.411 & 0.994 & 0.582 & 0.6971 & 718 & 1027 & 6 & 4 \\
C5 & 0.16 & 0.468 & 0.433 & 0.95 & 0.595 & 0.6901 & 686 & 897 & 136 & 36 \\
C6 & 0.25 & 0.455 & 0.427 & 0.949 & 0.589 & 0.6715 & 685 & 919 & 114 & 37 \\
C7 & 0.23 & 0.446 & 0.422 & 0.945 & 0.584 & 0.6728 & 682 & 933 & 100 & 40 \\
B1 & 0.2 & 0.436 & 0.421 & 0.992 & 0.591 & 0.6682 & 716 & 983 & 50 & 6 \\
B2 & 0.12 & 0.46 & 0.428 & 0.928 & 0.586 & 0.7409 & 670 & 895 & 138 & 52 \\
B3 & 0.1 & 0.437 & 0.418 & 0.938 & 0.578 & 0.8557 & 677 & 943 & 90 & 45 \\
B4 & 0.28 & 0.411 & 0.411 & 0.999 & 0.583 & 0.686 & 721 & 1032 & 1 & 1 \\
\bottomrule
\end{tabular}%
}
\end{minipage}

\end{document}